\documentclass[11pt]{article}

\usepackage[T1]{fontenc}
\usepackage[utf8]{inputenc}
\usepackage{times}
\usepackage{microtype}
\usepackage[letterpaper,margin=1in]{geometry}
\usepackage{amsmath,amssymb}
\usepackage{booktabs,multirow,makecell,array}
\usepackage{graphicx}
\usepackage[table]{xcolor}
\usepackage{tikz}
\usepackage{tabularx}
\usepackage{booktabs}
\usetikzlibrary{arrows.meta,positioning,fit,calc}
\usepackage{enumitem}
\usepackage{caption}
\usepackage{fancyhdr}
\usepackage{url}
\usepackage[numbers,sort&compress]{natbib}
\usepackage{hyperref}
\hypersetup{colorlinks=true,linkcolor=blue,citecolor=blue,urlcolor=blue}
\usepackage{cleveref}
\usepackage{placeins}

\hypersetup{colorlinks=true, allcolors=blue}
\definecolor{linkblue}{HTML}{1A6FB8}
\hypersetup{colorlinks=true, urlcolor=linkblue, linkcolor=linkblue}

\definecolor{headerblue}{RGB}{219,229,241}
\definecolor{f1tan}{RGB}{253,233,217}

\newcommand{\hflogo}{\raisebox{-0.18ex}{\includegraphics[height=1em]{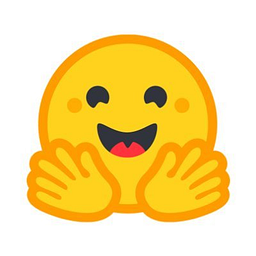}}}
\setlist{itemsep=2pt,topsep=3pt}
\newcommand{\hg}{HaloGuard 2.0}
\newcommand{\hgs}{HaloGuard 2.0-0.8B}
\newcommand{\hgp}{HaloGuard 2.0-Poly-0.8B}

\newcommand{\headeright}{A Preprint}
\newcommand{\shorttitle}{\textit{\hg: Open-Weights Constitutional Output Guard}}
\fancypagestyle{plain}{\fancyhf{}\fancyfoot[C]{\thepage}}

\begin{document}
\thispagestyle{plain}

\vspace*{0.3in}
\noindent\rule{\textwidth}{2pt}
\begin{center}
  {\LARGE \scshape HaloGuard 2.0: An Open-Weights\\[3pt]
   Constitutional Output Guard for \\[3pt] Multilingual Response Safety
   \par}
\end{center}
\vspace{-6pt}
\noindent\rule{\textwidth}{2pt}

\begin{center}
  {\scshape Technical Report}\\[16pt]
    ~\href{https://orcid.org/0000-0002-1310-3442}{\includegraphics[scale=0.06]{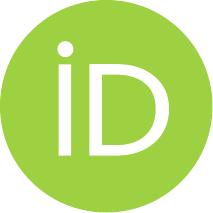}}\textbf{Ashmiya Lenin}$^{1,2*}$\quad
  \textbf{Navaneeth Sangameswaran}$^{1}$\quad
  \textbf{Preetham S}$^{1}$\\[4pt]
  $^{1}$Astroware AI, Delaware, USA\\
  $^{2}$Karunya Institute of Technology and Sciences, Coimbatore, India\\
  \text{ashmiya@astroware.ai} $|$ \text{nav@astroware.ai} $|$ \text{preetham@astroware.ai}\\
  $^{*}$Correspondence: ashmiya@astroware.ai\\[6pt]
   {\small\ttfamily
  	\hflogo\href{https://huggingface.co/collections/astroware/haloguard-10}{https://huggingface.co/collections/astroware/haloguard20}}%
\end{center}

\begin{center}{\large\scshape Abstract}\end{center}
\vspace{-4pt}
\begin{quote}
\small
An input guard determines whether a request should be forwarded to the model; however, it does not observe the model's subsequent output. Harm that emerges during generation, either due to a jailbreak bypassing the input layer or a benign request
shifting into operational detail, becomes observable exclusively on the output side and only as tokens are generated.
This work introduces HaloGuard 2.0, an open-weight streaming output guard that assigns a safe/unsafe probability to each
response token in a single forward pass, enabling deployment systems to terminate the stream once harm is detected.
The guard reuses the constitution of HaloGuard 1.0, the input guard, to organise a 478,532-response
training corpus, and supervises each token beginning at the point where the response transitions to being harmful.
We find that a guard trained on conventionally structured data develops a shortcut specific to streaming: it evaluates the
prompt rather than the response, and correctly handles only 13.9\% of matched safe and unsafe response pairs. We correct this with a second training phase on 106,043 counterfactual responses built around the base guard’s own failure modes, and release the equal-weight average of the base and corrective models. On six English response-safety benchmarks, HaloGuard 2.0-0.8B reaches a macro F1 of 83.6 (83.3 on held-out halves of the decision-rule search), up from 72.8, with 38\% fewer false positives. It matches the strongest full-response guards with 5–10× fewer parameters and is 2.6 points above the strongest streaming guard, Qwen3Guard-Stream-4B. A multilingual variant, HaloGuard 2.0-Poly-0.8B, trained on PolyGuardMix, reaches 83.1 F1 on the response labels of PolyGuardPrompts across 17 languages, 5.0 points above the strongest baseline. On one H100, the guard costs 0.37 ms per token per stream when moderating 64 concurrent responses. Both guards are hardened continuously through the weekly red-teaming protocol of HaloGuard 1.0, extended here to the output layer.
\end{quote}
\section{Introduction}
\label{sec:intro}

Guard models now sit on both sides of most production LLM deployments. On the input side, a classifier inspects the user's request before generation. On the output side, a second classifier inspects what the model produced. Our previous release, HaloGuard~1.0~\citep{halo1}, addressed the first of these layers. This report addresses the second, because screening the input is not enough on its own. Some harmful responses come from requests that look acceptable in isolation: a multi-turn conversation that escalates slowly, a role-play that drifts, or a jailbreak whose payload only resolves once the model starts answering. In agentic and retrieval-augmented settings, the model's context also contains text the user never wrote, including tool outputs, web pages and documents, any of which may carry an injected instruction. Aligned models still fail: a benign question about pesticide toxicity can produce a response that turns, several paragraphs in, into a synthesis route. Prompt-only scoring cannot see this.
Typically, moderation occurs after response generation is complete. This approach is suitable for batch auditing but inadequate for
interactive applications where responses are streamed to users incrementally. By the time a post-hoc classifier
returns a verdict, potentially harmful content may already be visible to the user. An alternative is to buffer the
entire response prior to release, which eliminates the latency advantage of streaming. A streaming guard avoids
both drawbacks by incrementally scoring the response and halting generation once the content exceeds the policy
boundary. Streaming makes the classification problem harder in a specific way. A response prefix is often
ambiguous. For example, “To synthesise this compound, you would first need” may continue as a refusal, a textbook
overview, or a detailed procedure. A guard that triggers on the prefix may interrupt legitimate educational
responses, while a guard that waits for certainty may allow operational content to be delivered. The decision is no longer solely
between safe and unsafe content. It also concerns the point at which evidence becomes sufficient. Streaming introduces two additional errors to
the standard false positive (FP) and false negative (FN) errors: triggering on a prefix that would have remained safe, and triggering
only after unsafe content has already been presented to the user.

The fundamental challenge remains the same as that which motivated the development of HaloGuard 1.0. Responses concerning weapons, malware,
self-harm, drugs, or extremism are frequently legitimate. For instance, a security engineer may inquire about how a class of exploit works
to facilitate patching, a nurse may ask about overdose thresholds, or a historian may inquire about chemical attacks in the
First World War. Both harmful and harmless versions of such responses share similar topics and vocabulary. What
distinguishes them is operational specificity, actionability, and the direction of intent. A guard that learns the
vocabulary rather than the boundary interrupts responses in precisely those domains where users most require a capable
model. Streaming introduces a second shortcut that does not occur on the input side. In typical
response corpora nearly every harmful response is harmful from its first token, so a
guard can score well by judging the prompt and never reading the answer. We test for it directly
with matched pairs, in which one prompt has a safe and an unsafe completion and only
the response can tell them apart (\Cref{sec:streaming}). A guard trained on
conventionally shaped data separates few of these pairs, and this measurement is what
motivates the corrective second phase of our recipe. We introduce \textbf{\hg{}}, a constitutional response classifier that scores model
outputs as they stream. It reuses the HaloGuard~1.0 constitution as the organising
structure of its training corpus and carries the input-side techniques, boundary hard
negatives and paired counterfactuals, over to responses. The English model, \hgs{}, is built on Qwen3.5-0.8B with a token-level
classification head. Its training data comprises 584,575 English responses in two
phases: 478,532 constitution-organised responses in Phase~1 and 106,043 corrective
counterfactuals in Phase~2, with every unsafe response supervised token by token from
the point where harm begins. For multilingual coverage we train a second variant,
\hgp{}, on 1,055,192 response-labelled conversations from PolyGuardMix~\citep{polyguard}
in 17 languages. Our English corpus, 584,575 responses, is about half the size of
Qwen3Guard's 1.19M samples~\citep{qwen3guard} and about 1\% of Shieldstral's
54.1M~\citep{shieldstral} (\Cref{tab:data-compare}). We evaluate on six English response-safety benchmarks,
HarmBench (HarmB), SafeRLHF, BeaverTails, XSTest-Response (XSTest-R), Aegis~2.0 and
WildGuardTest (WildG), and on the response-harmfulness labels of PolyGuardPrompts for
multilingual transfer. We compare against the open guards evaluated by
Qwen3Guard~\citep{qwen3guard} and Shieldstral~\citep{shieldstral}: LlamaGuard3-8B~\citep{llama3},
LlamaGuard4-12B~\citep{llamaguard4}, WildGuard-7B~\citep{wildguard}, ShieldGemma-9B and
-27B~\citep{shieldgemma}, NemoGuard-8B~\citep{aegis2}, PolyGuard-Qwen-7B~\citep{polyguard}, OmniGuard-7B~\citep{omniguard}, Nemotron-3.5-Safety-4B~\citep{nemotron35safety} and GPT-OSS-Safeguard-20B~\citep{gptosssafeguard}, as
well as Qwen3Guard-Gen and Qwen3Guard-Stream at 0.6B, 4B and 8B and Shieldstral-3B
itself. We report precision, recall, F1, FP rate (FPR) and
FN rate (FNR).

This report describes how we built a streaming output guard that reads the response. Our contributions are:

\begin{enumerate}
\item \textbf{An open constitutional output guard.} \hgs{} is a Qwen3.5-0.8B
  backbone with a linear token head that emits a safe/unsafe probability for every
  response token in a single forward pass. Its training corpus is organised by the HaloGuard 1.0 constitution, so input and output verdicts for an interaction refer to the same policies (\Cref{sec:constitution}).
\item \textbf{Onset-derived token supervision.} Each unsafe training response carries an onset, the first character of the span that makes it unsafe. Tokens before the onset are labelled safe and tokens from it onward unsafe, and tokens near it are up-weighted in the loss, so the guard learns when to fire, not only whether to.
\item \textbf{A two-phase method against the prompt shortcut.} We introduce a
matched-pair test for the shortcut and add it as a third FP mode
beside the boundary and baseline modes of HaloGuard~1.0. A corrective second
phase on counterfactual responses, merged into the base guard by weight
averaging, raises six-benchmark macro F1 from 72.8 to 83.6 (83.3 held out) and
removes 38\% of FPs.
\item \textbf{A multilingual variant.} \hgp{}, a second 0.8B model with the same
  architecture and objective, is trained on 1,055,192 response-labelled
  conversations from PolyGuardMix in 17 languages~\citep{polyguard} and reaches
  83.1 F1 on the response labels of PolyGuardPrompts, against 78.1 for the
  strongest baseline (Qwen3Guard-Gen-4B) and 75.8 for the strongest streaming guard
  (Qwen3Guard-Stream-8B).
\item \textbf{Continuous red-teaming of both layers.} The weekly red-teaming
  protocol of HaloGuard~1.0 now attacks a stack in which both guards are active, and
  attributes each successful attack to the layer that should have stopped it
  (\Cref{app:redteam}).
\end{enumerate}

\Cref{sec:related} reviews related work. \Cref{sec:task} defines the streaming
task and the onset. \Cref{sec:data} describes the constitution on the response side
and the training data.
\Cref{sec:model} covers the model, training and serving, and \Cref{sec:eval}
the evaluation. \Cref{sec:limits} discusses limitations, and
\Cref{sec:conclusion} concludes. \Cref{app:per-language} gives per-language results,
and \Cref{app:redteam} describes the continuous red-teaming protocol.

\section{Related Work}
\label{sec:related}
LlamaGuard framed moderation as instruction-following classification over a risk taxonomy and applied one model to both prompts and responses~\citep{llamaguard}. WildGuard introduced refusal detection as an additional output,
which is significant for response moderation because a refusal and a harmful completion may share the same prompt but require
opposite labels~\citep{wildguard}.
ShieldGemma, NemoGuard and PolyGuard extended open moderation to new model
families, enterprise taxonomies and 17 languages~\citep{shieldgemma,aegis2,polyguard}. More recent guards add reasoning or inference-time policies: OmniGuard reasons before
labelling~\citep{omniguard}, while GPT-OSS-Safeguard, Nemotron-3.5-Safety and
Shieldstral accept a deployer-written policy at inference
time~\citep{gptosssafeguard,nemotron35safety,shieldstral}.
All of them classify a completed response, so they must wait for generation to
finish before returning a verdict. Constitutional Classifiers (CCs) trained a streaming
output classifier that scores each token of a response and stops generation when
the harm estimate crosses a threshold~\citep{constclassifiers}. The classifier
guards a frontier model and its weights are not public. Qwen3Guard-Stream attaches token-level classification heads to a Qwen3 backbone and
derives token labels by locating the first unsafe token of each unsafe response:
continuations are sampled from every prefix, scored by Qwen3Guard-Gen, and the
earliest prefix whose continuations are mostly unsafe is confirmed by an LLM
judge~\citep{qwen3guard}. \hg{} is closest in shape to Qwen3Guard-Stream and closest
in data philosophy to CCs. It differs in how onsets are
obtained (in the corrective corpus they are constructed at generation time, so they
are exact rather than estimated), in how they are used (up-weighted in the loss and
paired with matched safe responses), and in how the model is trained (two phases,
merged by weight averaging).

Constitutional AI showed
that written principles can supervise model behaviour~\citep{constai}, and
CCs used a constitution of restricted and permitted content
to synthesise classifier training data~\citep{constclassifiers}. HaloGuard~1.0
scaled this to a full safety constitution (\Cref{sec:constitution}) and paired every harmful prompt with a harmless
counterfactual~\citep{halo1}. Counterfactually augmented data~\citep{kaushik} and
contrast sets~\citep{contrastsets} harden decision boundaries with minimally
different examples of opposite label. For responses, the natural minimal pair is
two completions to the same prompt, one harmful and one not. Averaging the weights of fine-tuned models from a
shared initialisation, or interpolating a fine-tuned model with its starting point,
often outperforms either endpoint and recovers robustness that fine-tuning
lost~\citep{wiseft,modelsoups}. We use the simplest form, a uniform average of the
base guard and its corrective fine-tune, and report what it gains and loses for a
streaming output classifier. HarmB provides harmful behaviours and
judged completions~\citep{harmbench}. BeaverTails and SafeRLHF provide
question--answer pairs with human safety labels~\citep{beavertails,saferlhf}.
WildG includes response-harmfulness labels on adversarial and in-the-wild
interactions~\citep{wildguard}, Aegis~2.0 provides response labels under the NVIDIA
content-safety taxonomy~\citep{aegis2}, and XSTest-R labels responses to
XSTest's benign-but-risky-sounding prompts~\citep{xstest,wildguard}.
\Cref{tab:positioning} positions \hg{} against prior response guards.

\begin{table}[!ht]
	\centering
	\small
	\setlength{\tabcolsep}{4pt}
\begin{tabularx}{\textwidth}{@{}
		>{\raggedright\arraybackslash}p{2.9cm}
		>{\raggedright\arraybackslash}X
		>{\raggedright\arraybackslash}p{2.3cm}
		>{\raggedright\arraybackslash}p{1.7cm}
		>{\raggedright\arraybackslash}X@{}}
		\toprule
			\textbf{Model} & 	\textbf{Training data} & 	\textbf{Scored} & 	\textbf{Streaming} & 	\textbf{Token supervision} \\
		\midrule
		LlamaGuard~\citep{llamaguard}       & collected + annotated  & full response & no & -- \\
		WildGuard~\citep{wildguard}         & collected + synthetic  & full response & no & -- \\
		ShieldGemma~\citep{shieldgemma}     & synthetic              & full response & no & -- \\
		PolyGuard~\citep{polyguard}         & collected + translated & full response & no & -- \\
		Qwen3Guard-Stream~\citep{qwen3guard} & collected + synthetic & every token & yes, open & first unsafe token, from rollouts + LLM judge \\
		CCs~\citep{constclassifiers} & constitution $\to$ synthetic & every token & yes, closed & prefixes trained toward the full-response label \\
		Shieldstral~\citep{shieldstral} & collected + synthetic (54.1M) & full response & no & -- \\
			\midrule
		\textbf{\hg{} (ours)} & \textbf{constitution $\to$ synthetic (English); PolyGuardMix (multilingual)} & \textbf{every response token} & \textbf{yes, open} & \textbf{onset constructed at generation (Phase~2)} \\
		\bottomrule
	\end{tabularx}
	\caption{Positioning of \hg{} relative to prior output guards.}
	\label{tab:positioning}
\end{table}
\section{Task Definition and Scope}
\label{sec:task}

\hg{} reads the user prompt $x$ together with the response $r=(r_1,\dots,r_N)$ and
emits, at every response position $t$, the probability that the response
\emph{so far} is unsafe:
\begin{equation}
p_t = p_\theta(\textsc{unsafe} \mid x, r_{\le t}), \qquad t = 1,\dots,N.
\label{eq:pt}
\end{equation}
Prompt positions are never scored. The prompt is context: whether ``Sure, here is
the recipe'' is harmful depends entirely on what was asked. A probability per token is not yet a decision. A single
threshold on $p_t$ fires on isolated spikes; requiring a long run of high scores
cannot fire on a short harmful response at all. We use a two-arm rule:
\begin{equation}
t^\star = \min\bigl\{\, t \;:\; p_t \ge \tau_{\max} \;\;\lor\;\; \min(p_{t-1}, p_t) \ge \tau_{\mathrm{run}} \bigr\},
\label{eq:rule}
\end{equation}
with $\tau_{\max}=0.975$ and $\tau_{\mathrm{run}}=0.70$. The first arm catches a
single high-confidence token, which is what a short compliance looks like (``Sure.
Step 1: \dots''). The second arm catches sustained moderate evidence, which is what
gradual drift looks like. The second arm is the two-token debounce that Qwen3Guard-Stream applies to every
flag~\citep{qwen3guard}; we pair it with a higher single-token arm so that very
short harmful responses can still fire. The verdict is latched: once the rule fires, the
response is unsafe and stays unsafe. If no such $t$ exists the response completes
and is safe. For benchmark scoring, the streaming decision collapses to a binary
verdict: \textsc{unsafe} if and only if $t^\star$ exists. For the English model we hold every candidate rule to one structural constraint: it must be able to
fire on a harmful response of one to three tokens. Rules that require long runs or
window averages score well on benchmarks whose unsafe responses are long, but they
are blind to short harmful answers by construction, and that blindness is invisible
in aggregate F1 because such responses are rare in most test sets. The arms
of \Cref{eq:rule} together do satisfy it: the first can fire on a single token, the second from
the second token on. Each training response also carries one of three
behaviours. \emph{Complied}: the model answers the request. \emph{Refused}: it
declines without providing harmful content. \emph{Mixed}: it partly refuses and
partly complies. The safety label depends only on content. A response that opens
``I can't help with that, but here is how it is generally done'' and then gives a
procedure is unsafe; the opening refusal does not change the label, and the onset
occurs where the procedure begins.

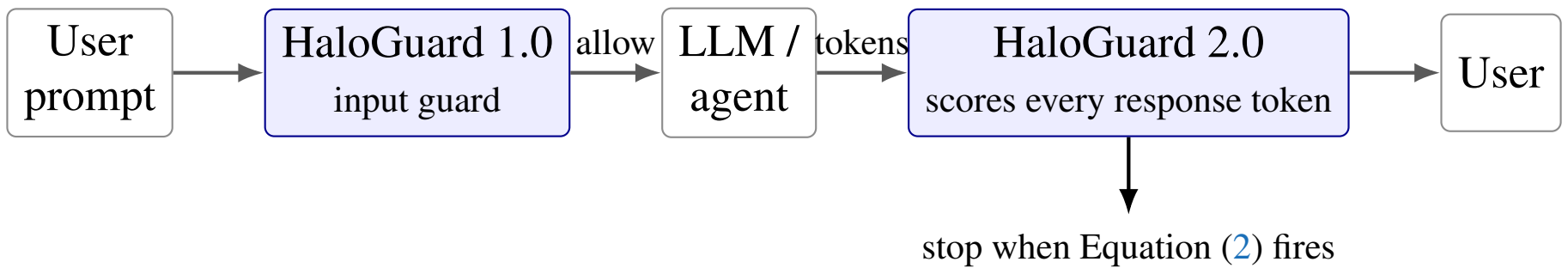
\begin{figure}[t]
\centering
\begin{tikzpicture}[
  node distance=5mm and 7mm,
  box/.style={draw=black!45, rounded corners=2pt, minimum height=9mm,
              align=center, font=\small, fill=white},
  guard/.style={box, fill=blue!7, draw=blue!55!black},
  arr/.style={-{Latex[length=2mm]}, thick, black!65}]
  \node[box] (u) {User\\prompt};
  \node[guard, right=of u] (g1) {HaloGuard 1.0\\\scriptsize input guard};
  \node[box, right=of g1] (llm) {LLM /\\agent};
  \node[guard, right=of llm, minimum width=30mm] (g2)
       {\hg\\\scriptsize scores every response token};
  \node[box, right=of g2] (out) {User};
  \draw[arr] (u) -- (g1);
  \draw[arr] (g1) -- node[above, font=\scriptsize,text=black!70!black]{allow} (llm);
  \draw[arr] (llm) -- node[above, font=\scriptsize,text=black!70!black]{tokens} (g2);
  \draw[arr] (g2) -- (out);
  \draw[arr, black!70!black] (g2.south) -- ++(0,-6mm)
       node[below, font=\scriptsize, text=black!70!black]{stop when \Cref{eq:rule} fires};
\end{tikzpicture}
\caption{HaloGuard~1.0 screens the request; \hg{} scores the response as it is
generated and stops the stream once harm is confirmed. Both guards are built from
the same constitution.}
\label{fig:deployment}
\end{figure}

\hg{} sits on the output side of an LLM application
(\Cref{fig:deployment}). It provides a signal; the application decides the action,
such as stopping the stream, replacing the remainder with a safe completion, or
letting the response finish and routing it to review. In scope are direct harmful
completions, legitimate responses on sensitive topics, and responses where harm
appears after a disclaimer, refusal, role-play frame or long benign lead-in. Tool
calls, actions and retrieved documents are out of scope except as text that appears
in the response. The response side needs one concept the input side does not: the \emph{onset}, the
first character of the earliest span that makes the response unsafe under its
policy. A response that restates the question, adds a caveat and then gives a
procedure has its onset at the procedure. A response that opens with the procedure
has its onset at zero. The onset turns one response-level label into a label for
every prefix (\Cref{eq:labels}).

\section{Constitution and Training Data}
\label{sec:data}

Training data comes in three corpora: one for each training phase of the English model
(\Cref{tab:data}), and PolyGuardMix for the multilingual variant
(\Cref{sec:ml-data}). The Phase~1 corpus is organised by the constitution and is the
response-side counterpart of the HaloGuard~1.0 corpus. The Phase~2 corpus is built
after Phase~1, from an analysis of the base guard's errors. For reference, the HaloGuard~1.0 prompt corpus has 1,259,451 records: 35.0\%
harmful, 47.9\% harmless-boundary and 17.2\% shared-harmless~\citep{halo1}. The
Phase~1 response corpus draws more heavily on the harmful bucket (55.2\% of
responses, \Cref{tab:data}), because a harmful prompt yields refusals and safe
answers as well as unsafe ones: only 30.2\% of Phase~1 responses are unsafe.

\begin{table}[!ht]
\centering
\small
\begin{tabular}{@{}llrr@{}}
\toprule
\textbf{Corpus} & \textbf{Slice} & \textbf{Responses} & \textbf{Share} \\
\midrule
\multirow{4}{*}{\makecell[l]{Phase 1\\(base)}}
  & Harmful pool              & 264,059 & 55.2\% \\
  & Harmless-boundary pool    & 180,301 & 37.7\% \\
  & Shared-harmless pool      &  34,172 &  7.1\% \\
  & \textit{Total}            & \textit{478,532} & \\
\midrule
\multirow{3}{*}{\makecell[l]{Phase 1,\\by behaviour}}
  & Complied                  & 350,646 & 73.3\% \\
  & Refused                   &  94,455 & 19.7\% \\
  & Mixed                     &  33,431 &  7.0\% \\
\midrule
\multirow{3}{*}{\makecell[l]{Phase 2\\(corrective)}}
  & Safe                      &  52,761 & 49.8\% \\
  & Unsafe                    &  53,282 & 50.2\% \\
  & \textit{Total}            & \textit{106,043} & \\
\midrule
\multicolumn{2}{@{}l}{Unsafe responses with onset after token 0, Phase 1} & 3,122$^{\dagger}$ & 2.2\% \\
\multicolumn{2}{@{}l}{Unsafe responses with onset after token 0, Phase 2} & 27,280 & 51.2\% \\
\bottomrule
\end{tabular}
\caption{Training data. Pools follow the HaloGuard~1.0 constitution;
$^{\dagger}$Of 144,721 unsafe Phase~1 responses, 141,599 are harmful from the
first token. Pool and behaviour shares are over all Phase~1 responses.}
\label{tab:data}
\end{table}

\subsection{The Constitution on the Response Side}
\label{sec:constitution}

\hg{} uses the HaloGuard~1.0 constitution without modification: 46 policies (29
harmful and 17 shared-harmless), 490 categories and 2,940 subcategories, with the
four risk tiers and three safety buckets defined there~\citep{halo1}. Sharing one
constitution across both layers has a practical benefit. An input verdict and an output verdict for the same interaction reference the same policy, enabling the deployer to compare outcomes and audit any disagreements. This subsection outlines the changes that occur when the constitution organises responses rather than prompts.

\textbf{From harmful requests to harmful content.} On the input side, a subcategory identifies a harmful request. On the output side, it identifies harmful content, defined as text whose presence renders a response unsafe. These categories can diverge in both directions. For example, a request for arson techniques is inherently harmful, but a response may be safe if it consists of a refusal or an explanation of fire investigation methods. Conversely, an innocuous request may result in harmful content if the model over-complies. Therefore, response labels are assigned exclusively to the response and are not inherited from the prompt.

\textbf{Buckets on the Response side.} The three buckets of HaloGuard~1.0 carry
over with response-level meanings. Harmful responses contain content the
constitution marks as harmful. Harmless-boundary responses address the topic
of a harmful subcategory and use its vocabulary, but stay at a level of detail the
constitution permits, for example a conceptual, defensive, clinical or historical
treatment. Shared-harmless responses answer ordinary requests and anchor the
guard's prior on routine traffic. HaloGuard~1.0's \textit{information\_depth}
coverage axis does the work on the response side that it did on the prompt side:
it spreads the harmless-boundary responses across levels of depth so that the guard
learns where on that axis each policy's boundary sits, rather than learning that a
topic is unsafe.

\subsection{Phase 1: The Constitutional Response Corpus}
\label{sec:phase1-data}

\begin{figure}[!ht]
\centering
\includegraphics[width=0.62\linewidth]{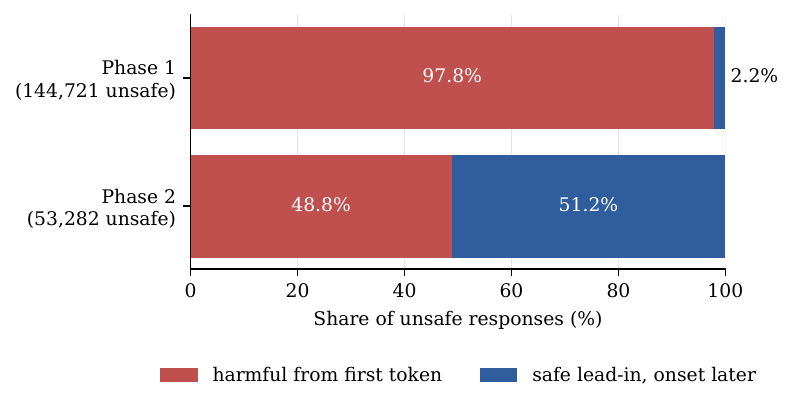}
\caption{Almost every unsafe response in the Phase~1 corpus is
harmful from its first token; half of the Phase~2 corpus turns harmful
mid-response.}
\label{fig:onset-skew}
\end{figure}

The Phase~1 corpus contains 478,532 prompt--response pairs drawn from the three
buckets of the constitution. Prompts come from the HaloGuard~1.0 prompt pool, so
every response inherits its prompt's policy, category, subcategory and bucket.
Responses were produced by a pool of generator models with different alignment
profiles: aligned models supply natural refusals, partial compliance and hedged
answers; models with reduced safety tuning supply compliant harmful answers. Using
several generators keeps the guard from keying on any one model's writing style. Each response has a safety label (safe, unsafe) and, if unsafe, an onset. The Phase~1 corpus has a property that turns out to
matter more than its size: almost every unsafe response is unsafe from its first
token (\Cref{fig:onset-skew}). Only 3,122 of 144,721 unsafe responses (2.2\%) have
any safe text before the harm. This is the natural shape of harmful completions
from compliant generators, which tend to start answering immediately. It is also
the condition under which a guard can minimise its loss by deciding from the prompt
(\Cref{sec:streaming}).

\subsection{Three FP Modes}
\label{sec:fp-modes}

HaloGuard~1.0 distinguished two kinds of FP. A boundary FP flags
a legitimate response near a policy edge for its topic or vocabulary, such as a
defensive explanation of an exploit class. A baseline FP flags an ordinary
response for an incidental feature. On responses the baseline mode is mostly
lexical: the guard fires on ``crack'' meaning a pastry, ``coke'' meaning the drink,
``shoot'' meaning photography, or ``kill the lights''. Streaming adds a third mode, which we call the \emph{prompt-driven} FP: a safe
response is flagged because of the prompt it answers. The guard activates at or near the initial response token, prior to any substantive content in the response, and it triggers identically for refusals, educational answers, and harmful responses. Prompt-driven FPs resemble over-refusal in aggregate metrics; however, their underlying causes and corresponding solutions differ. Increasing harmless-boundary data
does not eliminate these FPs; only safe responses paired with an unsafe response to the same prompt remove them. \Cref{tab:fp-modes} maps each mode to the data that targets it.

\begin{table}[!ht]
\centering
\small
\begin{tabular}{@{}p{2.4cm}p{5.6cm}p{6.4cm}@{}}
\toprule
\textbf{Mode} & \textbf{Failure pattern} & \textbf{Data response} \\
\midrule
Boundary & A legitimate response near a policy edge is flagged for its topic or vocabulary. & Harmless-boundary pool spread across information depth (Phase~1); safe twins of harmful responses (Phase~2). \\
Baseline (lexical) & An ordinary response is flagged for a harm-adjacent word used innocently. & Shared-harmless pool (Phase~1); lexical-collision responses with innocuous referents (Phase~2). \\
Prompt-driven & A safe response to a harmful-sounding prompt is flagged before it says anything harmful. & Matched safe/unsafe responses to the same prompt; unsafe responses with a safe lead-in and a late onset (Phase~2). \\
\bottomrule
\end{tabular}
\caption{The three FP modes targeted by \hg{}. The third is specific to
streaming output guards.}
\label{tab:fp-modes}
\end{table}

\subsection{Phase 2: The Corrective Corpus}
\label{sec:phase2-data}

The Phase~2 corpus is built from what the base output guard gets wrong. We collected errors from our internal development split and the six evaluation benchmarks, adjudicated these errors against the constitution (\Cref{tab:error-taxonomy}), removed benchmark mislabels and policy disagreements, and categorised the remaining genuine errors into failure archetypes: prompt-driven firing, lexical collisions, responses that initially refuse and then comply, subtle compliant harm in everyday-harm policies (property crime, privacy, drugs), and short responses.
We converted each archetype into a set of generation seeds, and expanded each seed into multiple new responses. Generation was governed by four rules.

\begin{table}[!ht]
\centering
\small
\begin{tabular}{@{}lr@{}}
\toprule
\textbf{Error source} & \textbf{Share of 3,107 errors} \\
\midrule
Genuine model error        & 66.6\% \\
Benchmark mislabel         & 19.4\% \\
Policy disagreement        & 18.1\% \\
\bottomrule
\end{tabular}
\caption{Adjudicated benchmark errors for an intermediate model. The shares exceed 100\%  because certain errors are attributed to two causes.}
\label{tab:error-taxonomy}
\end{table}

\begin{enumerate}[leftmargin=*]
\item \textbf{Every safe response has an unsafe twin.} The unsafe twin addresses the same prompt, utilises similar vocabulary and length, and differs solely in the presence or absence of uplift. An initial experiment that relabelled 19,033 borderline responses as safe, without including matched unsafe examples, halved the FPR but cost 17 points of recall, indicating that the guard model began classifying the entire topic as safe.
\item \textbf{Both twins come from the same generator.} Otherwise, the generator’s stylistic features may serve as a proxy for labelling. In an early corrective corpus where twins were sourced from different generators, matched-pair accuracy (\Cref{sec:streaming}) on the held-out set increased from 13.9\%  to 88\% . However, the six-benchmark F1 score improved by only 0.16. We attribute part of this gap to the generator confound.
\item \textbf{Onset is constructed, not inferred.} Approximately half of unsafe responses (51.2\%) begin with a safe segment, such as a restatement, caveat, refusal, or overview, and subsequently transition to harmful content. The safe span and the harmful continuation are generated as
  separate pieces, so the onset is exact by construction.
\item \textbf{Every pair is checked.} An independent judge verifies both labels before a pair is included in the corpus. Each response is compared to a hash blocklist containing all 12,347 benchmark evaluation rows and is also screened for near-duplicates.
\end{enumerate}

\subsection{Multilingual Data}
\label{sec:ml-data}
The multilingual variant adds PolyGuardMix~\citep{polyguard}. Of its 1,910,372
rows, 830,180 contain no response and are dropped; of the remaining 1,080,192, we
hold out 5,000 for validation and 20,000 as a development pool and train on
1,055,192. Responses are 96.6\% unique across roughly 660K prompts, so many prompts
carry several responses, which suits an output guard. PolyGuardMix carries
response-level labels only. We
verified that none of the 29,325 prompt--response pairs of PolyGuardPrompts (29,053 of which carry a response label) appears
in PolyGuardMix. \Cref{tab:data-totals} summarises the three training corpora. \Cref{tab:data-compare} compares this training data with that of the two closest open guards, Qwen3Guard and Shieldstral.
\begin{table}[t]
	\centering
	\small
	\begin{tabular}{@{}llrrrr@{}}
		\toprule
		\textbf{Corpus} & \textbf{Language} & \textbf{Responses} & \textbf{Unsafe}& \textbf{Unsafe share} & \textbf{Onset $>0$} \\
		\midrule
		Phase 1: Constitutional       & English & 478,532   & 144,721 & 30.2\% & 2.2\% \\
		Phase 2: Corrective-constitutional           & English & 106,043   &  53,282 & 50.2\% & 51.2\% \\
		\midrule
		\text{English total}        &         & \text{584,575} & \text{198,003} & \text{33.9\%} & \\
		\midrule
		PolyGuardMix (multilingual)   & 17      & 1,055,192 & 251,791 & 23.9\% & n/a \\
		\midrule
		\textbf{Total}                &         & \textbf{1,639,767} & 449,794 & 27.4\% & \\
		\bottomrule
	\end{tabular}
	\caption{Training data by corpus and language. Onset $>0$ is the share of unsafe responses that open with
		safe text before the harm. PolyGuardMix contributes the 1,080,192 of its 1,910,372
		rows that contain a response; 5,000 are held out for validation and 20,000 as a
		development pool, leaving 1,055,192 for training. It shares no prompt--response pair
		with PolyGuardPrompts.}
	\label{tab:data-totals}
\end{table}

\begin{table}[!ht]
	\centering
	\small
	\resizebox{\textwidth}{!}{%
		\begin{tabular}{@{}lllll@{}}
			\toprule
			\textbf{Model} & \textbf{Sizes} & \textbf{Training data} & \textbf{Languages} & \textbf{Streaming} \\
			\midrule
			Qwen3Guard-Gen~\citep{qwen3guard}    & 0.6B, 4B, 8B & $>$1.19M (58.8\% responses) & 119 & no \\
			Qwen3Guard-Stream~\citep{qwen3guard} & 0.6B, 4B, 8B & derived from Gen pool & 119 & token-level \\
			Shieldstral~\citep{shieldstral}      & 3B           & 54.1M (45.2M text, 4.4M synthetic, 4.5M image) & multilingual & no \\
			\textbf{\hgs{}}                      & \textbf{0.8B} & \textbf{584.6K En} & \textbf{English} & \textbf{token-level} \\
			\textbf{\hgp{}}                      & \textbf{0.8B} & \textbf{PolyGuardMix + corrective} & \textbf{17} & \textbf{token-level} \\
			\bottomrule
	\end{tabular}}
	\caption{Training data and scope relative to the two closest open guards.}
	\label{tab:data-compare}
\end{table}
\section{Model, Training and Serving}
\label{sec:model}

\subsection{Architecture}
\label{sec:arch}

The guard's input is a short policy preamble, followed by the user prompt and the
response. The preamble is a one-paragraph definition of safe and unsafe responses
(refusals, caveats, redirection and academic or defensive framing are safe;
operational uplift is unsafe, including a response that refuses first and then
complies) and the list of 27 constitution categories. HaloGuard~1.0 is a generative classifier: it reads the whole prompt and then writes
a label. That interface does not suit streaming, because producing a label at every
response token would multiply inference cost by the label length. \hg{} instead
places a linear head on the final hidden state of a Qwen3.5-0.8B
backbone~\citep{qwen35}:
\begin{equation}
p_t = \operatorname{softmax}\!\bigl(W h_t + b\bigr)_{\textsc{unsafe}}, \qquad W \in \mathbb{R}^{2\times 1024},
\end{equation}
where $h_t$ is the final-layer state at response position $t$. Because the backbone
is causal, $h_t$ depends only on $x$ and $r_{\le t}$, so one forward pass yields a
verdict for every prefix. The vision tower and language-model head of the base
checkpoint are dropped. Two details of this setup mattered in practice. First, the head is initialised to
zero. Qwen3.5's final hidden states have a root-mean-square of about 4.2, so the
usual $\mathcal{N}(0, 0.02^2)$ initialisation produces logits with a standard
deviation of about 2.7 before any training. In our runs that led to a first-step gradient
norm above 500 and a collapse to a constant prediction. With a zero head, training
starts at a loss of exactly $\ln 2$. Second, the backbone is hybrid: 18 of its 24 layers
are Gated DeltaNet linear-attention layers~\citep{gateddeltanet} with a recurrent
state, and 6 are full attention. For streaming inference this is an advantage,
because the 18 recurrent layers carry a fixed-size state from token to token. For
training it rules out sequence packing (\Cref{sec:training}).
\subsection{Token Labels and Loss}
\label{sec:labels}

The onset is stored as a character offset and mapped to the first token whose span
ends after it. A token that straddles the onset is therefore labelled unsafe. For a
response with onset token $o$,
\begin{equation}
y_t = \begin{cases}
0 & \text{safe response, or } t < o,\\
1 & \text{unsafe response and } t \ge o,\\
\varnothing & \text{prompt position, contested response, or unsafe response whose harmful span was truncated.}
\end{cases}
\label{eq:labels}
\end{equation}
Positions labelled $\varnothing$ are excluded from the loss. There is no masked
tolerance band after the onset: the first unsafe token is supervised as unsafe.

The loss is a weighted token-level cross-entropy over supervised positions $\mathcal{T}$:
\begin{equation}
\mathcal{L} = \frac{\sum_{t\in\mathcal{T}} w_t \,\mathrm{CE}(y_t, p_t)}{\sum_{t\in\mathcal{T}} w_t},
\qquad
w_t = \begin{cases} \lambda & \text{if } |t - o| \le 16,\\ 1 & \text{otherwise,}\end{cases}
\label{eq:loss}
\end{equation}
where the up-weighting applies only to responses with a safe-to-unsafe transition
($0 < o \le N$). A response that is unsafe from its first token has no transition in
the response and receives uniform weight. The weight $\lambda$ concentrates gradient
on the few tokens that decide \emph{when} the guard fires; we use $\lambda=5$ in
Phase~1 and $\lambda=3$ in Phase~2, where transitions are 23 times more frequent.

\subsection{Two-Phase Training}
\label{sec:training}

\begin{figure}[!ht]
\centering
\includegraphics[width=\textwidth]{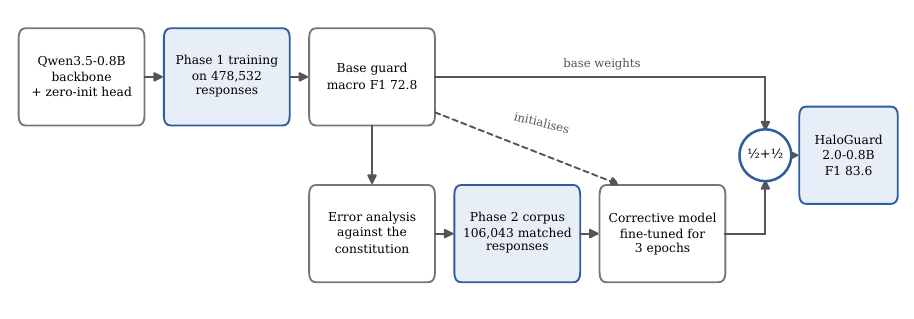}
\caption{Two-phase training of \hgs{}. Phase~1 trains a base guard on the
constitution-organised response corpus. Phase~2 analyses the base guard's errors,
builds a corrective corpus of matched responses, fine-tunes a copy of the base
guard on it, and releases the equal-weight average of the two models (the
$\frac{1}{2}+\frac{1}{2}$ node, \Cref{eq:average}).}
\label{fig:training-phases}
\end{figure}

\paragraph{Phase 1: Base training.} The backbone and head are trained jointly on
the 478,532-response corpus for two epochs. The base guard reaches a macro F1 of
72.8 on the six benchmarks and, as \Cref{sec:streaming} shows, learns largely
to judge the prompt.
\Cref{fig:training-phases} summarises the recipe and \Cref{tab:config} gives the
configuration.

\paragraph{Phase 2: Corrective refinement.} A copy of the base guard is fine-tuned
on the 106,043-response corrective corpus alone, for three epochs, with the same
objective. The released model averages the two sets of weights:
\begin{equation}
\theta_{\text{released}} = \tfrac{1}{2}\,\theta_{\text{base}} + \tfrac{1}{2}\,\theta_{\text{corrective}},
\label{eq:average}
\end{equation}
applied to every backbone and head parameter. Averaging is not a cosmetic step. The corrective model on its own scores below the base guard on the benchmarks, and so does retraining from scratch
on the corrective data mixed into the Phase~1 corpus (69.1 and 70.7 respectively,
against 72.8 for the base guard, measured in an earlier Phase~2 round). The
corrective model's errors are not the base guard's errors: it fixes the prompt
shortcut and the lexical triggers but compresses its scores, so that no single
threshold separates the classes as well as before. Averaging keeps most of the
correction and restores the score scale.

\textbf{Packing.} We train without sequence packing. In the hybrid backbone,
FlashAttention-2's variable-length kernel~\citep{fa2} isolates packed examples in the six
full-attention layers, but the recurrent state of the linear-attention layers runs
across the boundary between packed examples. On a two-layer probe, the same example
produced hidden states differing by up to 0.23 depending on what was packed before
it. Training is unpacked, with length-grouped batching to limit padding.
\paragraph{Multilingual variant.} \hgp{} uses the same
architecture, objective and decision-rule form. It is trained on the PolyGuardMix
corpus (\Cref{sec:ml-data}) from Qwen3.5-0.8B for two epochs (65,950 steps,
6.9~h on 2$\times$H100); benchmark F1 plateaued after about 40,000 steps.
The PolyGuardMix-trained base is then refined with corrective supervision mined
from its benchmark failures, following the same recipe as the English model, and
the result is averaged with the base at equal weight, as in
\Cref{eq:average}. It uses a single run arm: the guard fires when 3 consecutive
response tokens are all $\ge 0.675$, latched. Unlike \Cref{eq:rule}, this rule
cannot fire on a response shorter than three tokens.

\begin{table}[!ht]
\centering
\small
\begin{tabular}{@{}lll@{}}
\toprule
\textbf{Setting} & \textbf{Phase 1: Base Training} & \textbf{Phase 2: Corrective fine-tuning} \\
\midrule
Initialisation        & Qwen3.5-0.8B; head zero-init & Phase~1 base guard \\
Head                  & linear, $1024\!\to\!2$ & inherited \\
Training responses    & 478,532 & 106,043 \\
Epochs / steps        & 2 / 29,910 & 3 / 9,942 (released: step 9,000) \\ 
Learning rate         & $3\!\times\!10^{-5}$, head $10\times$ & $3\!\times\!10^{-5}$, head $10\times$; 3\% warm-up, linear decay \\
Effective batch       & 32 (8 $\times$ 2 accum.\ $\times$ 2 GPUs) & 32 \\
Onset weight $\lambda$ / window & 5.0 / $\pm16$ tokens & 3.0 / $\pm16$ tokens \\
Max length            & 2,048 tokens, unpacked & 2,048 tokens, unpacked \\
Precision / parallelism & bf16, FlashAttention-2, DDP & same \\
Hardware / wall clock & 2$\times$H100 80GB / 4.6 h & 2$\times$H100 80GB / 62 min \\
\midrule
Decision rule  & \multicolumn{2}{l}{$\max_t p_t \ge 0.975$ or two consecutive $p_t \ge 0.70$, latched (\Cref{eq:rule})} \\
\bottomrule
\end{tabular}
\caption{Training and serving configuration of \hgs{}.}
\label{tab:config}
\end{table}

\subsection{Serving}
\label{sec:serving}

At inference the guard reads the policy preamble and the prompt once, then receives response tokens as the generator emits them. Each new token advances the recurrent state of the linear-attention layers and the key–value cache of the full-attention layers, and produces one new $p_t$. The rule of eq. (2) is applied after every token. On a single idle H100 with fused linear-attention kernels (flash-linear-attention 0.5.0, causal-conv1d 1.7.0, PyTorch 2.11), and a 512-token prefill of preamble and prompt, the incremental step costs 22.2 ms at batch size 1 and 23.7 ms at batch size 64, or 0.37 ms per token per stream when moderating 64 concurrent responses: 2,705 response tokens per second in aggregate. The step is constant in response length: 22.6, 22.2 and 22.8 ms measured 128, 512 and 1,024 response tokens into the cache, roughly 1,536 tokens of total context at the deepest point, because the eighteen recurrent layers carry a fixed-size state and only the six full-attention layers grow theirs. At batch 1 the step is dominated by kernel-launch overhead rather than computation — scaling from batch 1 to 64 costs 6.7\% more wall time — so single-stream latency would fall substantially under CUDA-graph capture, which we do not apply here. Because the guard scores token $t$ while the generator produces $t{+}1$, it adds latency only when its step exceeds the generator's; in deployment, where one guard serves many concurrent streams, the batch-64 figure is the operative one. The guard was trained on inputs of up to 2,048 tokens. For longer responses we score two windows, one anchored at the start of the response and one at the end, and take the element-wise maximum where they overlap. Scoring only the head window lets an attacker bury a payload after a long filler; scoring only the tail lets them bury it at the start. With both windows, the guard covers roughly twice its training length.

\section{Evaluation}
\label{sec:eval}

We evaluate response harmfulness on six English benchmarks (\Cref{tab:benchmarks})
and on the response labels of PolyGuardPrompts. For each benchmark we use the
response-level safety label; refusal labels are not used.

\begin{table}[!ht]
\centering
\small
\begin{tabular}{@{}lrrl@{}}
\toprule
\textbf{Benchmark} & \textbf{Responses} & \textbf{Unsafe} & \textbf{Notes} \\
\midrule
HarmBench~\citep{harmbench}          &   400 & 50.0\% & judged completions to harmful behaviours \\
SafeRLHF~\citep{saferlhf}            & 5,978 & 55.2\% & test split, response safety \\
BeaverTails~\citep{beavertails}      & 3,021 & 57.4\% & test split, QA pairs \\
WildGuardTest~\citep{wildguard}      & 1,689 & 16.7\% & response harmfulness \\
Aegis~2.0~\citep{aegis2}             &   813 & 48.5\% & response-labelled test split \\
XSTest-Response~\citep{xstest,wildguard} & 446 & 17.5\% & responses to XSTest prompts \\
\midrule
\text{Total}                       & \text{12,347} & & \\
\midrule
PolyGuardPrompts~\citep{polyguard} (response) & 29,053 & 16.6\% & 1,709 conversations $\times$ 17 languages \\
\bottomrule
\end{tabular}
\caption{Response-safety evaluation suite.}
\label{tab:benchmarks}
\end{table}

\hg{} is scored by its streaming verdict: a response is unsafe
if \Cref{eq:rule} fires at any response token. Only response tokens are scored. This
is stricter than full-response classification because the guard commits before it
sees the end. One decision rule is used for all benchmarks; we do not tune
thresholds per benchmark, because a deployed guard cannot know which distribution a
request comes from, and in our measurements per-benchmark thresholds inflate macro
F1 by about three points. The two thresholds of \Cref{eq:rule}
were selected by a grid search (step 0.025) on the benchmarks themselves. To
estimate the optimism this introduces, we split every benchmark in half, chose the
rule on one half and scored it on the other, for two random seeds in both
directions. The held-out macro F1 is 83.3 (83.5 and 83.2 for the two seeds),
against 83.6 in-sample.

\subsection{Results}
\label{sec:results}

\Cref{tab:family} places the two HaloGuard guards side by side, in the format
Qwen3Guard uses to summarise its family~\citep{qwen3guard}: HaloGuard~1.0 on
prompts and \hg{} on responses, in English and across 17 languages.
\Cref{tab:main} then reports response-harmfulness F1 per benchmark against open
baselines and \Cref{tab:per-bench-errors} the error rates behind it, and
\Cref{tab:halo1-prompt} gives the input-layer results of HaloGuard~1.0; the operating
points of both released models are compared in \Cref{sec:multilingual}.

\begin{table}[!ht]
	\centering
	\small
	\setlength{\tabcolsep}{4pt}
	\begin{tabular}{@{}lrcccc@{}}
		\toprule
		& & \multicolumn{2}{c}{English} & \multicolumn{2}{c}{Multilingual (PolyGuardPrompts)} \\
		\cmidrule(lr){3-4}\cmidrule(lr){5-6}
		\textbf{Model} & Size & Prompt & Response & Prompt & Response \\
		\midrule
		LlamaGuard3         & 8B   & 79.4 & 70.5 & 68.0 & 65.7 \\
		LlamaGuard4         & 12B  & 75.9 & 68.9 & 62.1$^{\S}$ & 53.4 \\
		ShieldGemma         & 27B  & 70.0 & 65.6 & --   & 50.4 \\
		WildGuard           & 7B   & 85.8 & 81.4 & 73.0$^{\S}$ & 63.5 \\
		NemoGuard           & 8B   & 82.9 & 78.1 & --   & 69.5 \\
		PolyGuard-Smol      & 0.5B & --   & --   & 83.8 & 66.8 \\
		PolyGuard-Ministral & 8B   & --   & --   & 86.0 & 73.8 \\
		PolyGuard-Qwen      & 7B   & 87.0 & 72.9 & 87.1 & 74.1 \\
		Qwen3Guard-Gen      & 0.6B & 84.7 & 81.3 & --   & 74.2 \\
		Qwen3Guard-Gen      & 4B   & 85.6 & 83.6 & --   & 78.1 \\
		Qwen3Guard-Gen      & 8B   & 86.2 & 83.6 & 84.3$^{\S}$ & 77.6 \\
		Qwen3Guard-Stream   & 0.6B & 84.7 & 78.8 & --   & 70.6 \\
		Qwen3Guard-Stream   & 4B   & 86.2 & 81.0 & --   & 75.5 \\
		Qwen3Guard-Stream   & 8B   & 86.7 & 80.4 & --   & 75.8 \\
		Shieldstral         & 3B   & 88.9$^{\ddagger}$ & 82.5 & 84.6$^{\S}$ & 78.3$^{\S}$ \\
		GPT-OSS-Safeguard   & 20B  & 88.1$^{\ddagger}$ & 80.9 & 83.0$^{\S}$ & 80.0$^{\S}$ \\
		\midrule
		\textbf{HaloGuard 1.0}                 & 0.8B & 90.9 & --   & 86.1 & -- \\
		\textbf{HaloGuard 1.0}                 & 4B   & \textbf{92.1} & --   & \textbf{88.0} & -- \\
		\textbf{HaloGuard 2.0}                        & 0.8B & --   & \textbf{83.6} & --   & -- \\
		\textbf{HaloGuard 2.0-Poly}                        & 0.8B & --   & --   & --   & \textbf{83.1} \\
		\bottomrule
	\end{tabular}
	\caption{The HaloGuard family across tasks (F1). English prompt: average over the
		seven prompt benchmarks of HaloGuard~1.0 (OAI, Aegis, Aegis~2.0, ToxicChat,
		SimpleSafetyTests, HarmBench, WildGuardTest). English response: average over the six
		benchmarks of \Cref{tab:main}. Multilingual: PolyGuardPrompts, 17 languages. --
		marks a task the model does not perform or a score no source reports. Bold marks the
		best score per column; in the English response column \hgs{} leads at two decimals
		(83.63 against 83.57 for Qwen3Guard-Gen-4B and -8B, \Cref{tab:main}) with
		5--10$\times$ fewer parameters. English baselines as reported in \citet{halo1} and
		the Qwen3Guard report~\citep{qwen3guard} (Tables~2, 3 and 11; Qwen3Guard in strict
		mode). Multilingual baselines as reported in the PolyGuard paper~\citep{polyguard}
		(Table~1) and the Qwen3Guard report (Tables~6 and 15), which agree to within 0.1
		where they overlap.
		$^{\ddagger}$Mean of the seven per-benchmark scores reported by \citet{shieldstral}
		(Figure~5); that paper's scores for WildGuard, LlamaGuard4 and Qwen3Guard-8B on the
		same benchmarks are within 0.5 of the values above.
		$^{\S}$PolyGuard F1 as reported by \citet{shieldstral} (Figure~5) on its own
		29,240-sample split, with Qwen3Guard averaged over its strict and loose modes, so
		indicative only (\Cref{sec:multilingual}). On prompts, PolyGuard-Qwen is the only
		model scored under both protocols, and the two differ by 1.2 (85.9 against 87.1);
		on responses, five models overlap (\Cref{sec:multilingual}).}
	\label{tab:family}
\end{table}

\begin{table*}[!ht]
\centering
\footnotesize
\setlength{\tabcolsep}{3.5pt}
\begin{tabular}{@{}lrccccccc@{}}
\toprule
\textbf{Model} & \textbf{Size} & \textbf{HarmB} & \textbf{SafeRLHF} & \textbf{BeaverTails} & \textbf{XSTest-R} & \textbf{Aegis~2.0} & \textbf{WildG} & \textbf{Avg}. \\
\midrule
LlamaGuard3$^{\dagger}$          & 8B   & 84.5 & 45.2 & 67.9 & 89.8 & 66.1 & 69.5 & 70.50 \\
LlamaGuard4$^{\dagger}$         & 12B  & 83.3 & 42.5 & 68.6 & 88.9 & 63.7 & 66.4 & 68.90 \\
WildGuard$^{\dagger}$            & 7B   & 86.3 & 64.2 & 84.4 & \textbf{94.7} & 83.2 & 75.4 & 81.37 \\
ShieldGemma$^{\dagger}$          & 9B   & 60.4 & 44.2 & 62.4 & 86.3 & 70.8 & 49.9 & 62.33 \\
ShieldGemma$^{\dagger}$         & 27B  & 62.9 & 52.6 & 67.6 & 83.0 & 74.9 & 52.4 & 65.57 \\
NemoGuard$^{\dagger}$            & 8B   & 81.4 & 57.6 & 78.5 & 86.2 & \textbf{87.6} & 77.5 & 78.13 \\
PolyGuard-Qwen$^{\dagger}$       & 7B   & 71.1 & 63.3 & 79.5 & 63.4 & 81.9 & 77.9 & 72.85 \\
OmniGuard$^{\ddagger}$           & 7B   & 79.9 & 64.2 & 83.3 & 85.1 & 85.5 & 72.1 & 78.35 \\
Nemotron-3.5-Safety$^{\ddagger}$ & 4B   & 85.3 & 63.2 & 83.3 & 86.9 & 84.9 & 77.6 & 80.20 \\
GPT-OSS-Safeguard$^{\ddagger}$  & 20B  & 88.2 & 63.8 & 83.8 & 93.8 & 75.2 & \textbf{80.7} & 80.92 \\
Shieldstral$^{\ddagger}$            & 3B   & 87.0 & 62.0 & 85.0 & 93.5 & 87.2 & 80.4 & 82.52 \\
Qwen3Guard-Gen$^{\dagger}$          & 0.6B & 85.0 & 66.6 & 86.1 & 89.7 & 84.2 & 76.3 & 81.32 \\
Qwen3Guard-Gen$^{\dagger}$          & 4B   & 86.7 & 69.8 & \textbf{86.6} & 92.7 & 86.1 & 79.5 & 83.57 \\
Qwen3Guard-Gen$^{\dagger}$          & 8B   & 87.2 & 70.5 & \textbf{86.6} & 92.1 & 86.1 & 78.9 & 83.57 \\
Qwen3Guard-Stream$^{\dagger}$       & 0.6B & 83.1 & 62.8 & 84.5 & 84.8 & 81.4 & 76.3 & 78.82 \\
Qwen3Guard-Stream$^{\dagger}$       & 4B   & 84.3 & 67.6 & 86.0 & 88.5 & 83.1 & 76.4 & 80.98 \\
Qwen3Guard-Stream$^{\dagger}$       & 8B   & 85.0 & 64.6 & 85.9 & 87.5 & 82.6 & 77.0 & 80.43 \\
\midrule
\hg{}, base guard (Phase 1)   & 0.8B & 91.6 & 71.7 & 72.3 & 68.0 & 74.1 & 58.9 & 72.76 \\
\textbf{\hgs{} (Ours)}    & \textbf{0.8B} & \textbf{96.4} & \textbf{81.7} & 83.0 & 85.4 & 83.5 & 71.9 & \textbf{83.63} \\
\bottomrule
\end{tabular}
\caption{\textbf{Response-harmfulness} F1 on six English benchmarks. \hg{} is scored by its
streaming verdict under one global decision rule; the released model's held-out
average is 83.3. The base guard is shown at its own best global rule. Best
score per column in bold. The Avg.\ column is given to two decimals because the
top three averages are within 0.06: \hgs{} (83.63, 0.8B) and Qwen3Guard-Gen-4B
and -8B (83.57). Baseline averages are the mean of the reported
per-benchmark scores.
$^{\dagger}$As reported by \citet{qwen3guard} (Qwen3Guard models in strict mode).
$^{\ddagger}$As reported by \citet{shieldstral}.
Qwen3Guard's Think column is not part of our suite.}
\label{tab:main}
\end{table*}

\begin{figure}[t]
\centering
\includegraphics[width=0.62\linewidth]{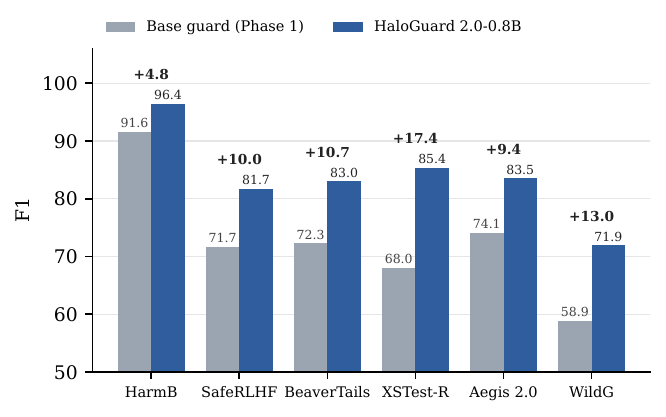}
\caption{F1 per benchmark, base guard vs.\ \hgs{}, with the change above
each pair. The largest gains are on the two benchmarks dominated by safe responses,
XSTest-R and WildG.}
\label{fig:per-benchmark}
\end{figure}

\begin{figure*}[!ht]
\centering
\includegraphics[width=\textwidth]{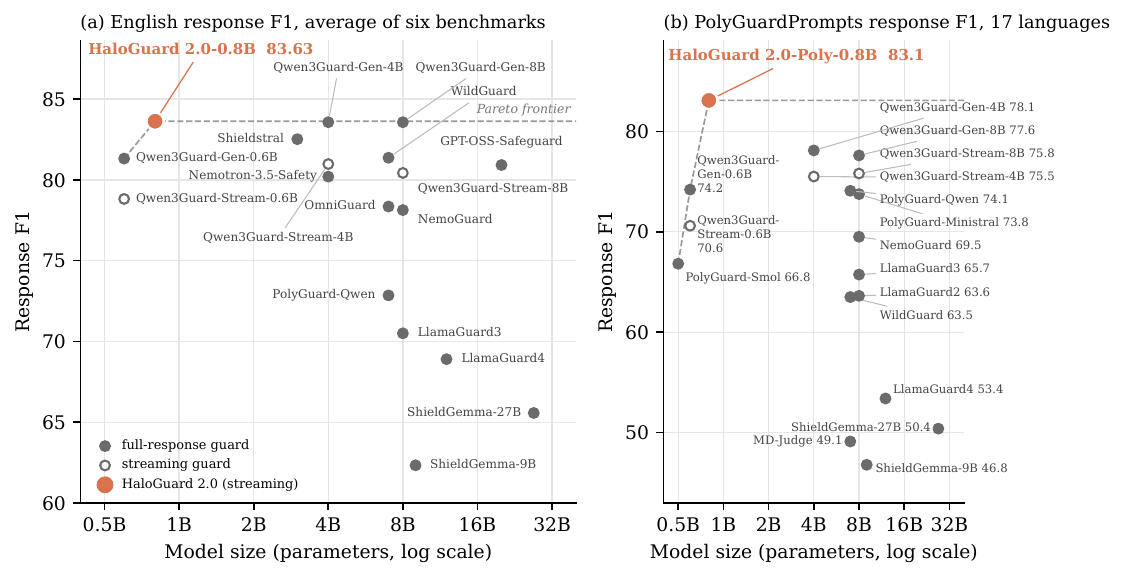}
\caption{Smaller model, higher average. Response F1 against model size (log
scale). (a) Average over the six English benchmarks of \Cref{tab:main}; \hgs{} has
the highest average (83.63, against 83.57 for Qwen3Guard-Gen-4B and -8B) at
5--10$\times$ smaller size, and is the highest-scoring streaming guard. Hollow markers are streaming
guards. (b) PolyGuardPrompts response F1 (\Cref{sec:multilingual}); baselines on the same
labels as reported by \citet{polyguard} and \citet{qwen3guard}. The dashed line is the Pareto frontier: no model is
both smaller and more accurate than a model on it.}
\label{fig:size-vs-f1}
\end{figure*}

Over the base guard, the largest gains (\Cref{fig:per-benchmark}) are on
XSTest-R (+17.4) and WildG (+13.0), the two benchmarks dominated by safe responses
to harmful-sounding prompts, where the prompt-driven and lexical FP modes
concentrate. FPs fall from 1,465 to 912.
Across the six benchmarks, \hgs{} has the highest average of any model in
\Cref{tab:main}, 83.63, ahead of Qwen3Guard-Gen-4B and -8B (83.57 each) with
5--10$\times$ fewer parameters. The margin over those two is small; the size gap
is not. \Cref{fig:size-vs-f1} plots these averages against model size, the
counterpart of the size plot in HaloGuard~1.0~\citep{halo1}. It is also above every streaming baseline: the best Qwen3Guard-Stream model, at 4B, averages 81.0, and the 0.6B
model, the closest in size, 78.8. The gap is not uniform. \hgs{} is the strongest
model on HarmB (96.4 against 88.2 for GPT-OSS-Safeguard-20B) and SafeRLHF (81.7
against 70.5 for Qwen3Guard-Gen-8B), and trails the best baseline on the other four,
by 3.6 points on BeaverTails, 4.1 on Aegis~2.0, 8.8 on WildG and 9.3 on
XSTest-R. Those two deficits differ in kind (\Cref{tab:per-bench-errors}). On
XSTest-R, where 82.5\% of responses are safe, the remaining errors are mostly false
positives: FPR 5.4 against FNR 6.4, or about 20 FPs against 5 FNs. On WildG they are
mostly misses: the FPR is 4.9 but the FNR is 30.1, about 69 FPs against 85 FNs, so
the guard misses close to a third of WildG's harmful responses. WildG remains the
weakest benchmark at 71.9 and is the main target of the next corrective round. On Aegis~2.0 the corrective phase bought recall alone: the FNR fell from 23.9 to 7.9 while the FPR barely moved, from 27.7 to 27.0. It is the one benchmark where the FP modes of \Cref{sec:fp-modes} did not respond to the corrective data. Pooled over all six benchmarks the released model misses more than it over-flags
(\Cref{tab:operating-point}). \Cref{tab:per-bench-errors} shows where: false
positives concentrate on Aegis~2.0, BeaverTails and SafeRLHF, misses on WildG,
SafeRLHF and BeaverTails.

\begin{table}[!ht]
\centering
\small
\begin{tabular}{@{}lcccc@{}}
\toprule
 & \multicolumn{2}{c}{\textbf{Base guard (Phase 1)}} & \multicolumn{2}{c}{\textbf{HaloGuard 2.0-0.8B}} \\
\cmidrule(lr){2-3}\cmidrule(lr){4-5}
\textbf{Benchmark} & FPR $\downarrow$ & FNR $\downarrow$ & FPR $\downarrow$ & FNR $\downarrow$ \\
\midrule
HarmB       & 0.0  & 15.5 & 0.0  & 7.0  \\
SafeRLHF    & 28.5 & 31.2 & 17.7 & 21.1 \\
BeaverTails & 26.0 & 32.5 & 18.3 & 19.4 \\
XSTest-R    & 14.1 & 14.1 & 5.4  & 6.4  \\
Aegis~2.0   & 27.7 & 23.9 & 27.0 & 7.9  \\
WildG       & 14.1 & 28.7 & 4.9  & 30.1 \\
\bottomrule
\end{tabular}
\caption{FPR and FNR (\%) per benchmark for the base guard and the released \hgs{},
each under its own decision rule (base guard: $\max_t p_t \ge 0.91$ or three
consecutive $p_t \ge 0.86$). Pooled rates for both are in \Cref{tab:operating-point}.}
\label{tab:per-bench-errors}
\end{table}

\paragraph{The input layer.} \Cref{tab:halo1-prompt} reproduces the prompt
results of HaloGuard~1.0~\citep{halo1}, the input guard that \hg{} is deployed
behind (\Cref{fig:deployment}), so that both layers of the stack are reported in
one place. On seven prompt-harmfulness benchmarks the 0.8B input guard averages
90.9 and the 4B model 92.1, the two highest averages among the open guards
compared. At its released operating point the 0.8B input guard has an FPR of 4.3
and an FNR of 9.5 (precision 91.8, recall 90.5)~\citep{halo1}. Three benchmarks
label both prompts and responses, and on each of them response classification is
the harder task for the same model size: HarmBench 98.7 on prompts against 96.4 on
responses, Aegis~2.0 87.9 against 83.5, and WildG 95.9 against 71.9. The
gap is largest on WildG, where adversarial prompts are easy to flag but
the harmful responses among them are hard to recall. Each guard is scored on
its own label, so this compares the two tasks, not the two models.

\begin{table*}[!ht]
\centering
\footnotesize
\setlength{\tabcolsep}{4.5pt}
\begin{tabular}{@{}lrcccccccc@{}}
\toprule
\textbf{Model} & \textbf{Size} & \textbf{OAI} & \textbf{Aegis} & \textbf{Aegis~2.0} & \textbf{ToxiC} & \textbf{SimpST} & \textbf{HarmB} & \textbf{WildG} & \textbf{Avg.} \\
\midrule
LlamaGuard4      & 12B  & 73.5 & 67.8 & 70.6 & 51.3 & 98.0 & 97.2 & 73.0 & 75.9 \\
WildGuard        & 7B   & 72.1 & 89.4 & 80.7 & 70.8 & 99.5 & 98.9 & 88.9 & 85.8 \\
ShieldGemma      & 27B  & 80.5 & 69.0 & 71.6 & 72.9 & 84.4 & 57.3 & 54.3 & 70.0 \\
NemoGuard        & 8B   & 81.0 & 81.4 & 86.8 & 75.6 & 98.5 & 75.2 & 81.6 & 82.9 \\
PolyGuard-Qwen   & 7B   & 74.1 & 90.3 & 86.3 & 71.5 & \textbf{100.0} & 98.7 & 88.1 & 87.0 \\
Qwen3Guard-Gen   & 0.6B & 66.5 & 90.8 & 85.0 & 65.1 & 99.0 & 98.7 & 87.7 & 84.7 \\
Qwen3Guard-Gen   & 4B   & 68.3 & 90.8 & 85.8 & 69.5 & 99.5 & \textbf{100.0} & 85.6 & 85.6 \\
Qwen3Guard-Gen   & 8B   & 68.8 & \textbf{91.4} & 86.1 & 68.9 & 99.5 & \textbf{100.0} & 88.9 & 86.2 \\
\midrule
\multirow{2}{*}{\textbf{HaloGuard 1.0}} & 0.8B & 83.7 & 86.7 & 87.9 & 83.5 & \textbf{100.0} & 98.7 & 95.9 & 90.9 \\
 & 4B & \textbf{87.4} & 88.0 & \textbf{89.2} & \textbf{84.5} & \textbf{100.0} & 99.2 & \textbf{96.2} & \textbf{92.1} \\
\bottomrule
\end{tabular}
\caption{Prompt harmfulness (F1) of the input guard, HaloGuard~1.0, across seven
safety benchmarks: OpenAI Moderation (OAI), Aegis, Aegis~2.0, ToxicChat (ToxiC),
SimpleSafetyTests (SimpST), HarmB and WildG. All values
as reported in \citet{halo1} (Table~14). Best per column in bold.}
\label{tab:halo1-prompt}
\end{table*}

\subsection{Streaming Behaviour}
\label{sec:streaming}

Benchmark F1 says whether the guard eventually reaches the right verdict. It does
not say whether it fires at the right time. For that we use a held-out onset set of
664 generated responses forming 332 matched pairs: one prompt, an unsafe response
whose harm begins mid-response, and a safe twin that never crosses. The set was
frozen before any Phase~2 data was generated and shares no prompts with training.
It was, however, produced by the same pipeline as the Phase~2 corpus, so it favours
models trained on that corpus; we read it alongside the benchmarks, not instead of
them. On this set the base guard handles only 13.9\% of pairs correctly, firing on the
unsafe response and not on its safe twin. It fires on
both in 45.8\% of pairs and on neither in 40.4\%. It never fires on the safe twin
alone. A guard that read the responses would at least sometimes get a pair
backwards; a guard that never does is deciding on what the twins share, which is
the prompt. Its median first fire on the unsafe responses comes 40 tokens
\emph{before} the onset. The same signature is visible on the benchmarks: in an
intermediate model, 37.4\% of FPs fired within the first three response
tokens, and the score at the first response token was 0.59 on responses the model
would falsely flag against 0.24 on those it would not, a gap that can only come from
the prompt. The corrective model learns to read the
response. Averaging gives part of that back. We averaged one corrective model with the
base guard at several weights, all under the same decision rule, in an earlier
Phase~2 round. Moving from the equal average to the corrective model alone
raises onset recall from 47.6\% to 72.6\% while macro F1 falls from 81.0 to 76.3,
about five points of onset recall for each point of F1.

\subsection{Multilingual Evaluation}
\label{sec:multilingual}

We evaluate multilingual transfer on PolyGuardPrompts~\citep{polyguard}, using its
response-harmfulness labels across 17 languages: 29,053
responses, 1,709 conversations each rendered in 17 languages. The benchmark is built
from translations of WildGuardMix and natural multilingual interactions. \hgp{}
is trained on PolyGuardMix (\Cref{sec:ml-data}), which shares none of the
benchmark's prompt--response pairs. Its corrective data is generated from seeds gathered from the benchmark’s failures (\Cref{sec:training}). \hgp{} reaches a response
F1 of 83.1, macro-averaged over the 17 languages (\Cref{tab:per-language}, \Cref{app:per-language}). That is
5.0 points above the strongest baseline on the same labels, Qwen3Guard-Gen-4B (78.1,
five times the size; \Cref{tab:polyguard}), and 7.3 points above the strongest streaming guard,
Qwen3Guard-Stream-8B (75.8). Baselines come from two reports, the PolyGuard
paper~\citep{polyguard} and the Qwen3Guard report~\citep{qwen3guard}; for the two
models both report, they agree to within 0.1 (LlamaGuard3 65.74 and 65.8,
PolyGuard-Qwen 74.08 and 74.0), so we treat them as one protocol. \Cref{tab:operating-point} sets the operating points of the two released models
side by side, and they have opposite error profiles. \hgs{} misses more than it
over-flags. \hgp{} over-flags more than it misses: its errors are 1,482 false
positives against 360 false negatives, so its remaining headroom is in precision,
not recall. Part of this is base rate. Only 16.6\% of PolyGuardPrompts responses are
unsafe, so even a low FPR produces more false positives than the guard has misses.

The corrective phase also works through a different error here. Over the
PolyGuardMix-only base guard it adds 18.2 points of F1, against 10.9 for the English
model (\Cref{tab:operating-point}). The English gain came mainly from fewer FPs, which fell by 38\%.
The multilingual gain comes mainly from fewer misses: FNs fall by 76\%,
from 1,531 to 360, while FPs fall by 26\%, from 1,999 to 1,482. The same
recipe works for both models, through a different mechanism in each. \Cref{tab:ml-groups} breaks the result down by language group and
\Cref{tab:per-language} (\Cref{app:per-language}) by language. The European and CJK
groups and Arabic stay within 4.4 points of English; performance falls for Thai
(73.9) and most sharply for Hindi (64.2). Recall drops most for Hindi: its FNR
reaches 22.2, against 4.6 for English, so the guard misses more than one harmful
response in five. Shieldstral~\citep{shieldstral} reports separate PolyGuard prompt and response F1
scores for ten guards (its Figure~5), under a different protocol: 29,240 samples
rather than the 29,053 response-labelled rows, and Qwen3Guard averaged over its
strict and loose modes. Five of its
models are also scored on the same labels above, and for four of them the two
figures are within 2 points (LlamaGuard4 54.6 against 53.4, WildGuard 65.3 against
63.5, PolyGuard-Qwen 74.8 against 74.1, Qwen3Guard-8B 78.1 against 77.6); the fifth,
ShieldGemma-9B, differs widely (31.8 against 46.8). We list Shieldstral's scores
only for the five models with no same-label score, in a separate block: the
strongest is GPT-OSS-Safeguard-20B at 80.0, and Shieldstral-3B scores 78.3. The
comparison with these rows is indicative rather than exact.

\begin{table}[!ht]
\centering
\small
\begin{tabular}{@{}lrc@{}}
\toprule
\textbf{Model} & \textbf{Size} & \textbf{Response F1} \\
\midrule
\multicolumn{3}{@{}l}{\textit{Same response labels}} \\
ShieldGemma$^{b}$             & 9B   & 46.8 \\
MD-Judge~\citep{mdjudge}$^{a}$ & 7B   & 49.12 \\
ShieldGemma$^{b}$             & 27B  & 50.4 \\
LlamaGuard4$^{b}$             & 12B  & 53.4 \\
WildGuard$^{b}$               & 7B   & 63.5 \\
LlamaGuard2~\citep{llamaguard2}$^{a}$ & 8B   & 63.62 \\
LlamaGuard3$^{a}$             & 8B   & 65.74 \\
PolyGuard-Smol$^{a}$          & 0.5B & 66.82 \\
NemoGuard$^{b}$               & 8B   & 69.5 \\
Qwen3Guard-Stream$^{b}$       & 0.6B & 70.6 \\
PolyGuard-Ministral$^{a}$     & 8B   & 73.75 \\
PolyGuard-Qwen$^{a}$          & 7B   & 74.08 \\
Qwen3Guard-Gen$^{b}$          & 0.6B & 74.2 \\
Qwen3Guard-Stream$^{b}$       & 4B   & 75.5 \\
Qwen3Guard-Stream$^{b}$       & 8B   & 75.8 \\
Qwen3Guard-Gen$^{b}$          & 8B   & 77.6 \\
Qwen3Guard-Gen$^{b}$          & 4B   & 78.1 \\
\midrule
\multicolumn{3}{@{}l}{\textit{Different protocol}$^{\S}$} \\
OmniGuard                     & 7B   & 69.1 \\
Nemotron-Safety~\citep{cultureguard} & 8B   & 73.7 \\
Nemotron-3.5-Safety           & 4B   & 75.3 \\
Shieldstral                   & 3B   & 78.3 \\
GPT-OSS-Safeguard             & 20B  & 80.0 \\
\midrule
\textbf{\hgp{}}     & \textbf{0.8B} & \textbf{83.1} \\
\bottomrule
\end{tabular}
\caption{PolyGuardPrompts response harmfulness F1 (17 languages), sorted by
score. Upper block: same response labels as \hgp{}; $^{a}$as reported in the
PolyGuard paper~\citep{polyguard} (Table~1), $^{b}$as reported in the Qwen3Guard
report~\citep{qwen3guard} (Tables~6 and 15; Qwen3Guard at its reported mode).
Scores are given at the precision each source reports. $^{\S}$Lower block: PolyGuard
scores as reported by \citet{shieldstral} (Figure~5, response panel) for models with no same-label
score, on 29,240 samples, so these rows are indicative only.}
\label{tab:polyguard}
\end{table}

\begin{table}[!ht]
\centering
\small
\begin{tabular}{@{}lcccc@{}}
\toprule
 & \multicolumn{2}{c}{\textbf{English} \footnotesize(6 benchmarks, 12,347)} & \multicolumn{2}{c}{\textbf{Multilingual} \footnotesize(PolyGuardPrompts, 29,053)} \\
\cmidrule(lr){2-3}\cmidrule(lr){4-5}
\textbf{Metric} & Base guard & \textbf{\hgs{}} & Base guard & \textbf{\hgp{}} \\
\midrule
Macro F1 $\uparrow$  & 72.8 & \textbf{83.6} & 64.9 & \textbf{83.1} \\
Precision $\uparrow$ & 74.0 & 84.1 & 62.4 & 75.5 \\
Recall $\uparrow$    & 69.8 & 80.5 & 68.3 & 92.5 \\
FPR $\downarrow$     & 23.0 & 14.3 & 8.3  & 6.1  \\
FNR $\downarrow$     & 30.2 & 19.5 & 31.7 & 7.5  \\
\bottomrule
\end{tabular}
\caption{Operating points of the two released models and their Phase~1 base guards,
each at its own decision rule. Macro F1 is averaged over the six benchmarks
(English) or the 17 languages (multilingual). Precision, recall, FPR and FNR are
pooled over all responses for English and macro-averaged over languages for the
multilingual models; PolyGuardPrompts is balanced across languages, so its pooled
and macro values differ by at most 0.4.}
\label{tab:operating-point}
\end{table}

\begin{table}[!ht]
\centering
\small
\begin{tabular}{@{}lccc@{}}
\toprule
\textbf{Language group} & \textbf{FPR $\downarrow$} & \textbf{FNR $\downarrow$} & \textbf{F1 $\uparrow$} \\
\midrule
English (reference)                  & 4.7  & 4.6  & 87.1 \\
CJK                                  & 6.0  & 7.5  & 83.1 \\
Indic                                & 12.8 & 22.2 & 64.2 \\
Arabic-script                        & 6.2  & 7.4  & 82.7 \\
European                             & 5.2  & 5.8  & 85.5 \\
Southeast Asian                      & 10.0 & 12.0 & 73.9 \\
\bottomrule
\end{tabular}
\caption{\hgp{} on PolyGuardPrompts by language group (\%). Groups follow
HaloGuard~1.0~\citep{halo1}: CJK is Chinese, Japanese and Korean; Indic is Hindi;
Arabic-script is Arabic; Southeast Asian is Thai; European is the
remaining ten languages. The macro average over all 17 languages is in
\Cref{tab:operating-point}; it is not the mean of these six rows.}
\label{tab:ml-groups}
\end{table}

\subsection{SOTA Status and Evaluation Scope}
\label{sec:sota}

\textbf{SOTA Status.} HaloGuard 2.0-0.8B is the strongest open streaming output guard in our comparison. Across six English response-safety benchmarks it reaches a macro F1 of 83.6 under one global streaming rule. That is 2.6 points above the best Qwen3Guard-Stream model (4B, 81.0) and 4.8 above the one closest in size (0.6B, 78.8). It is level with the strongest full-response guards, Qwen3Guard-Gen-4B and -8B (83.57), which see the complete response before deciding and have 5–10× more parameters. It is the best model on HarmBench (96.4) and on SafeRLHF (81.7, 11.2 points above the next model). Pooled over 12,347 responses it operates at 84.1 precision and 80.5 recall (FPR 14.3\% , FNR 19.5\% ), and its remaining error is mostly missed harm, concentrated on WildG. HaloGuard 2.0-Poly-0.8B reaches 83.1 response F1 on PolyGuardPrompts, macro-averaged over 17 languages. That is 5.0 points above Qwen3Guard-Gen-4B (78.1) and 7.3 above the strongest streaming guard, Qwen3Guard-Stream-8B (75.8), with a 6.1\%  FPR and 7.5\%  FNR. Its remaining headroom is in precision (75.5) rather than recall (92.5), and in Hindi and Thai. Together with HaloGuard 1.0, which leads the open input guards at 90.9 (0.8B) and 92.1 (4B) prompt F1, the 0.8B input and output guards form a complete moderation stack of 1.6B parameters.

\textbf{Evaluation scope.} Three evaluations reported by recent guards are not part of this release. RTP-LX, as in HaloGuard 1.0, measures the toxicity of stand-alone utterances (insults, slurs, identity attacks). Our constitution does not treat that as harmful response content: a response that is rude but gives no harmful uplift is safe under our policy. For the same reason we do not use Shieldstral’s per-language results, which pool RTP-LX with PolyGuard. Reasoning traces: Qwen3Guard also scores responses that include the model’s chain of thought. HaloGuard 2.0 was trained only on final responses. A trace of several hundred tokens that deliberates over a harmful request before declining it is a different input distribution, and covering it requires reasoning-trace training data, which we leave to a future release. Refusal detection: WildGuard and Qwen3Guard also label whether a response is a refusal. HaloGuard 2.0 labels harm only: a refusal is safe, and a response that refuses and then complies is unsafe from its onset. PolyGuardPrompts’ response labels, by contrast, match our task definition exactly. The gap between our English score on the benchmark (87.1) and the other 16 languages therefore measures how far the guard’s response boundary transfers across languages, which is the question the multilingual variant was built to answer.
\section{Limitations}
\label{sec:limits}
Every harmful
response that is caught has already emitted the tokens between its onset and the
fire. Deployers who cannot tolerate any leakage on the most severe policies should
hold back a short buffer of output before display. The averaging step that produces the benchmark gain costs mid-stream
detection (\Cref{sec:streaming}). A deployment that cares more about catching harm
early than about overall F1 may prefer a model weighted toward the corrective side. An earlier model in this line could be evaded by
prepending a single refusal sentence to a harmful response, and by placing a payload
after a long benign filler. The second is addressed in serving by dual-window
scoring (\Cref{sec:serving}); the first is addressed in training by unsafe
responses that refuse and then comply. Both attack classes are tracked in the
weekly red-teaming protocol (\Cref{app:redteam}), which is how new evasions reach
the corrective data. In the Phase~2 corpus onsets are exact by construction.
In the Phase~1 corpus they were inferred during corpus construction: a judge returned a verbatim quote of the first span that makes the response unsafe, and the quote was located in the response text by exact, normalised or prefix match to give the character offset. Responses on which two judges disagreed carry no onset, are marked contested, and are excluded from the loss (\Cref{eq:labels}). The boundary between analytical detail and a usable
procedure is sometimes genuinely contested. The multilingual variant has no onset
supervision outside English: PolyGuardMix carries response-level labels only, so every unsafe multilingual response is supervised as unsafe from its first token. The English model misses more than
it over-flags, most of all on WildG, while the multilingual model over-flags more
than it misses (\Cref{tab:operating-point}). A deployment should therefore calibrate
each model's threshold to its own error profile rather than reuse one operating
point across both. The multilingual rule needs
three tokens and cannot flag a one- or two-token harmful response.
Hindi and Thai
miss substantially more harmful content than the European and CJK groups
(\Cref{tab:ml-groups}). Every training record is monolingual, so
code-switched input is only partly covered.

\section{Conclusion}
\label{sec:conclusion}

HaloGuard 2.0 builds on HaloGuard 1.0 by shifting from per-prompt decisions to per-token decisions in the model’s output. The core rules stay the same, but now the system can look at both sides of each policy boundary, making it easier to see exactly where harm starts. The main idea is that a streaming guard needs to set two boundaries. The first is the content boundary, which works like input moderation and separates responses that encourage harm from those that just discuss harmful topics. The second is a temporal boundary, specific to streaming, that marks the point in a response where there is enough evidence to step in. If the guard is only trained on completions that are harmful from the start, it might miss this temporal boundary and focus only on the prompt. Comparing response pairs makes this issue clear. Adding a corrective training step with these pairs, and then averaging the model weights, raises the six-benchmark F1 score from 72.8 to 83.6. HaloGuard 2.0-Poly-0.8B (the multilingual version) uses the same training objective as PolyGuardMix and supports 17 languages, achieving an F1 of 83.1 on the PolyGuardPrompts response labels. Next up is finding training methods that maintain the streaming improvements during averaging. We also plan to try out larger models as the 0.8B version hasn't been tried above that parameter count yet. In the future we'd like to look at conversation level scoring, category attribution, tool call scoring and joint evaluation of input and output guards on shared test concepts.


\appendix

\section{Per-Language Results}
\label{app:per-language}

\Cref{tab:per-language} lists the response F1 of \hgp{} for each of the 17
languages of PolyGuardPrompts, sorted by score. The macro average is 83.1.

\begin{table}[!ht]
\centering
\small
\begin{tabular}{@{}lr@{\qquad}lr@{}}
\toprule
\textbf{Language} & \textbf{F1} & \textbf{Language} & \textbf{F1} \\
\midrule
Spanish    & 88.4 & Chinese  & 83.5 \\
English    & 87.1 & Japanese & 83.4 \\
Italian    & 86.6 & Russian  & 83.3 \\
Swedish    & 86.2 & Arabic   & 82.7 \\
French     & 86.0 & Korean   & 82.5 \\
Polish     & 85.9 & Czech    & 82.1 \\
Portuguese & 85.9 & Thai     & 73.9 \\
German     & 85.3 & Hindi    & 64.2 \\
Dutch      & 85.2 &          &      \\
\bottomrule
\end{tabular}
\caption{PolyGuardPrompts response F1 of \hgp{} per language.}
\label{tab:per-language}
\end{table}

\section{Continuous Adversarial Red-Teaming}
\label{app:redteam}

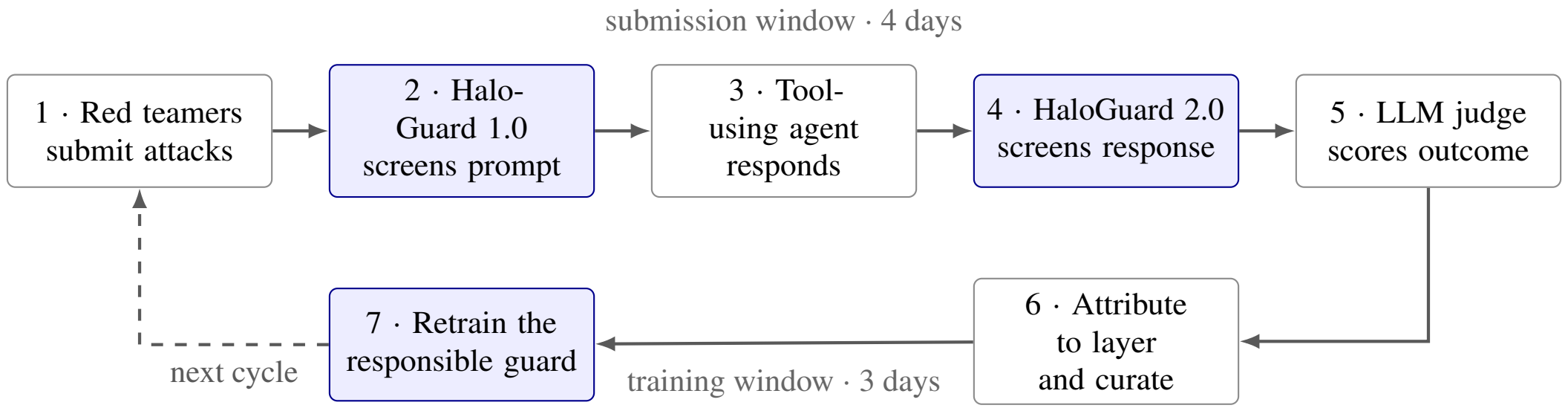
\begin{figure}[h]
\centering
\begin{tikzpicture}[
  node distance=4mm and 5mm,
  rtstep/.style={draw=black!45, rounded corners=2pt, minimum height=10mm,
               text width=21mm, align=center, font=\scriptsize, fill=white},
  guard/.style={rtstep, fill=blue!7, draw=blue!55!black},
  arr/.style={-{Latex[length=1.8mm]}, thick, black!65}]
  \node[rtstep]  (a) {1 $\cdot$ Red teamers\\submit attacks};
  \node[guard, right=of a] (b) {2 $\cdot$ HaloGuard 1.0\\screens prompt};
  \node[rtstep,  right=of b] (c) {3 $\cdot$ Tool-using agent\\responds};
  \node[guard, right=of c] (d) {4 $\cdot$ \hg\\screens response};
  \node[rtstep,  right=of d] (e) {5 $\cdot$ LLM judge\\scores outcome};
  \node[rtstep,  below=8mm of d] (f) {6 $\cdot$ Attribute to layer\\and curate};
  \node[guard, below=8mm of b] (g) {7 $\cdot$ Retrain the\\responsible guard};
  \draw[arr] (a) -- (b); \draw[arr] (b) -- (c); \draw[arr] (c) -- (d); \draw[arr] (d) -- (e);
  \draw[arr] (e.south) |- (f.east);
  \draw[arr] (f) -- (g);
  \draw[arr, dashed] (g.west) -| node[pos=0.25, below, font=\scriptsize]{next cycle} (a.south);
  \node[font=\scriptsize, text=black!60, above=1mm of c] {submission window $\cdot$ 4 days};
  \node[font=\scriptsize, text=black!60, below=1mm of $(f)!0.5!(g)$] {training window $\cdot$ 3 days};
\end{tikzpicture}
\caption{The continuous red-teaming loop with both guards active. Each successful
attack is attributed to the layer that should have stopped it and becomes training
data for that guard in the next cycle.}
\label{fig:redteam}
\end{figure}

Static benchmarks saturate once a defender adapts to them. HaloGuard~1.0 is therefore
hardened by an always-on red-teaming competition run in rolling seven-day cycles: a
four-day submission window in which an open pool of red teamers attacks a guarded,
tool-using agent, followed by a three-day training window in which successful attacks
are curated and used to retrain the guard~\citep{halo1}. We extend the same protocol
to the output layer (\Cref{fig:redteam}).

\textbf{Two guards, one attack.} Red teamers now attack a stack in which
HaloGuard~1.0 screens the prompt and \hg{} screens the response, so a successful
attack has to pass both layers. Each success is attributed to the layer that should
have caught it. If the prompt itself was harmful under the constitution and passed
the input guard, the attack becomes input-guard training data. If the prompt was
acceptable, or harmful in a way the input guard could not have recognised, and the
response turned harmful, the attack becomes output-guard training data. For the
output guard, the response is labelled at the token level like any other corrective
example: the curator marks the onset, and the attack enters the next Phase~2 corpus
with a matched safe response to the same prompt, so that the correction teaches the
guard what changed rather than which prompt to avoid.

\textbf{Output-specific attack classes.} Three attack classes target the output
layer and are tracked separately.
\emph{Late-onset} attacks have the model produce a long benign lead-in before the
harmful content, testing whether the guard keeps reading and whether its context
window can be exhausted.
\emph{Refuse-then-comply} attacks have the model open with a refusal or disclaimer and
then provide the content, testing whether an opening refusal suppresses the guard.
\emph{Format-shift} attacks have the model emit harmful content in an encoding, a
code block or another language, testing whether the guard follows the content across
surface changes.

\textbf{Scoring.} As in HaloGuard~1.0, each cycle covers two axes. Content attacks
are scored by an ensemble of LLM judges with a graded rubric (full success, partial
success, refusal); agentic attacks are scored by exact-string matching against
secrets planted in the agent's environment, which a judge cannot be talked into
accepting. Submissions are deduplicated with a similarity filter, and attacks aimed
at the judge rather than the agent are disallowed. 


\begin{thebibliography}{99}
\small

\bibitem[Sangameswaran et~al.(2026)]{halo1}
Navaneeth Sangameswaran, Preetham S, and Ashmiya Lenin.
\newblock {HaloGuard 1.0}: An open weights constitutional classifier for multilingual {AI} safety.
\newblock arXiv:2607.02079, 2026.

\bibitem[Kumar et~al.(2025)]{polyguard}
Priyanshu Kumar, Devansh Jain, Akhila Yerukola, Liwei Jiang, Himanshu Beniwal, Thomas Hartvigsen, and Maarten Sap.
\newblock {PolyGuard}: A multilingual safety moderation tool for 17 languages.
\newblock arXiv:2504.04377, 2025.

\bibitem[Qwen Team(2025)]{qwen3guard}
Qwen Team.
\newblock {Qwen3Guard} technical report.
\newblock arXiv:2510.14276, 2025.

\bibitem[Calvi et~al.(2026)]{shieldstral}
Antonia Calvi, Avinash Sooriyarachchi, Giada Pistilli, Guillaume Lample, Maarten Buyl, Maximilian Augustin, Maximilian M{\"u}ller, Pierre Stock, Tom Bewley, Wassim Bouaziz, Yimu Pan, et~al.
\newblock {Shieldstral}.
\newblock arXiv:2607.25857, 2026.

\bibitem[Chi et~al.(2024)]{llama3}  
Aaron Grattafiori, Abhimanyu Dubey, Abhinav Jauhri, et~al.
\newblock The {Llama 3} herd of models.
\newblock arXiv:2407.21783, 2024.

\bibitem[Meta(2025)]{llamaguard4}
Meta.
\newblock {Llama Guard 4}: Model card.
\newblock \url{https://developer.meta.com/ai/docs/model-cards-and-prompt-formats/llama-guard-4/}, 2025.

\bibitem[Han et~al.(2024)]{wildguard}
Seungju Han, Kavel Rao, Allyson Ettinger, Liwei Jiang, Bill~Yuchen Lin, Nathan Lambert, Yejin Choi, and Nouha Dziri.
\newblock {WildGuard}: Open one-stop moderation tools for safety risks, jailbreaks, and refusals of {LLMs}.
\newblock arXiv:2406.18495, 2024.

\bibitem[Zeng et~al.(2024)]{shieldgemma}
Wenjun Zeng, Yuchi Liu, Ryan Mullins, Ludovic Peran, et~al.
\newblock {ShieldGemma}: Generative {AI} content moderation based on {Gemma}.
\newblock arXiv:2407.21772, 2024.

\bibitem[Ghosh et~al.(2025)]{aegis2}
Shaona Ghosh, Prasoon Varshney, Makesh~Narsimhan Sreedhar, Aishwarya Padmakumar, Traian Rebedea, Jibin~Rajan Varghese, and Christopher Parisien.
\newblock {AEGIS2.0}: A diverse {AI} safety dataset and risks taxonomy for alignment of {LLM} guardrails.
\newblock In \emph{Proceedings of NAACL 2025 (Volume 1: Long Papers)}, pages 5992--6026, 2025.

\bibitem[Zhu et~al.(2025)]{omniguard}
Boyu Zhu, Xiaofei Wen, Wenjie~Jacky Mo, Tinghui Zhu, Yanan Xie, Peng Qi, and Muhao Chen.
\newblock {OmniGuard}: Unified omni-modal guardrails with deliberate reasoning.
\newblock arXiv:2512.02306, 2025.

\bibitem[Singh et~al.(2026)]{nemotron35safety}
Varun Singh, Isabel Hulseman, Anuj Doshi, and Shyamala Prayaga.
\newblock {Nemotron 3.5 Content Safety}: Customizable multimodal safety for global enterprise {AI}.
\newblock \url{https://huggingface.co/nvidia/Nemotron-3.5-Content-Safety}, 2026.

\bibitem[OpenAI(2025)]{gptosssafeguard}
OpenAI.
\newblock Introducing {GPT-OSS-Safeguard}.
\newblock \url{https://openai.com/index/introducing-gpt-oss-safeguard/}, 2025.

\bibitem[Inan et~al.(2023)]{llamaguard}
Hakan Inan, Kartikeya Upasani, Jianfeng Chi, Rashi Rungta, et~al.
\newblock {Llama Guard}: {LLM}-based input-output safeguard for human-{AI} conversations.
\newblock arXiv:2312.06674, 2023.

\bibitem[Sharma et~al.(2025)]{constclassifiers}
Mrinank Sharma, Meg Tong, Jesse Mu, Jerry Wei, et~al.
\newblock Constitutional classifiers: Defending against universal jailbreaks across thousands of hours of red teaming.
\newblock arXiv:2501.18837, 2025.

\bibitem[Bai et~al.(2022)]{constai}
Yuntao Bai, Saurav Kadavath, Sandipan Kundu, Amanda Askell, et~al.
\newblock Constitutional {AI}: Harmlessness from {AI} feedback.
\newblock arXiv:2212.08073, 2022.

\bibitem[Kaushik et~al.(2020)]{kaushik}
Divyansh Kaushik, Eduard Hovy, and Zachary~C. Lipton.
\newblock Learning the difference that makes a difference with counterfactually-augmented data.
\newblock In \emph{ICLR}, 2020.

\bibitem[Gardner et~al.(2020)]{contrastsets}
Matt Gardner, Yoav Artzi, Victoria Basmov, Jonathan Berant, et~al.
\newblock Evaluating models' local decision boundaries via contrast sets.
\newblock In \emph{Findings of EMNLP}, pages 1307--1323, 2020.

\bibitem[Wortsman et~al.(2022b)]{wiseft}
Mitchell Wortsman, Gabriel Ilharco, Jong~Wook Kim, Mike Li, Simon Kornblith, et~al.
\newblock Robust fine-tuning of zero-shot models.
\newblock In \emph{CVPR}, 2022.

\bibitem[Wortsman et~al.(2022a)]{modelsoups}
Mitchell Wortsman, Gabriel Ilharco, Samir~Yitzhak Gadre, Rebecca Roelofs, et~al.
\newblock Model soups: Averaging weights of multiple fine-tuned models improves accuracy without increasing inference time.
\newblock In \emph{ICML}, 2022.

\bibitem[Mazeika et~al.(2024)]{harmbench}
Mantas Mazeika, Long Phan, Xuwang Yin, Andy Zou, et~al.
\newblock {HarmBench}: A standardized evaluation framework for automated red teaming and robust refusal.
\newblock arXiv:2402.04249, 2024.

\bibitem[Ji et~al.(2023)]{beavertails}
Jiaming Ji, Mickel Liu, Juntao Dai, Xuehai Pan, Chi Zhang, Ce~Bian, Ruiyang Sun, Yizhou Wang, and Yaodong Yang.
\newblock {BeaverTails}: Towards improved safety alignment of {LLM} via a human-preference dataset.
\newblock arXiv:2307.04657, 2023.

\bibitem[Ji et~al.(2024)]{saferlhf}
Jiaming Ji, Donghai Hong, Borong Zhang, Boyuan Chen, et~al.
\newblock {PKU-SafeRLHF}: Towards multi-level safety alignment for {LLMs} with human preference.
\newblock arXiv:2406.15513, 2024.

\bibitem[R{\"o}ttger et~al.(2024)]{xstest}
Paul R{\"o}ttger, Hannah~Rose Kirk, Bertie Vidgen, Giuseppe Attanasio, Federico Bianchi, and Dirk Hovy.
\newblock {XSTest}: A test suite for identifying exaggerated safety behaviours in large language models.
\newblock arXiv:2308.01263, 2024.

\bibitem[Qwen Team(2026)]{qwen35}
Qwen Team.
\newblock {Qwen3.5}: Towards native multimodal agents.
\newblock \url{https://qwen.ai/blog?id=qwen3.5}, February 2026.

\bibitem[Yang et~al.(2025)]{gateddeltanet}
Songlin Yang, Jan Kautz, and Ali Hatamizadeh.
\newblock Gated delta networks: Improving {Mamba2} with delta rule.
\newblock In \emph{ICLR}, 2025.

\bibitem[Dao(2023)]{fa2}
Tri Dao.
\newblock {FlashAttention-2}: Faster attention with better parallelism and work partitioning.
\newblock arXiv:2307.08691, 2023.

\bibitem[Li et~al.(2024)]{mdjudge}
Lijun Li, Bowen Dong, Ruohui Wang, Xuhao Hu, Wangmeng Zuo, Dahua Lin, Yu~Qiao, and Jing Shao.
\newblock {SALAD-Bench}: A hierarchical and comprehensive safety benchmark for large language models.
\newblock arXiv:2402.05044, 2024.

\bibitem[Meta(2024)]{llamaguard2}
Llama Team, AI~@~Meta.
\newblock {Meta Llama Guard 2}: Model card.
\newblock \url{https://huggingface.co/meta-llama/Meta-Llama-Guard-2-8B}, 2024.

\bibitem[Joshi et~al.(2025)]{cultureguard}
Raviraj Joshi, Rakesh Paul, Kanishk Singla, Anusha Kamath, Michael Evans, Katherine Luna, Shaona Ghosh, Utkarsh Vaidya, Eileen Long, Sanjay~Singh Chauhan, and Niranjan Wartikar.
\newblock {CultureGuard}: Towards culturally-aware dataset and guard model for multilingual safety applications.
\newblock In \emph{Proceedings of IJCNLP-AACL 2025}, 2025.

\end{thebibliography}
\end{document}